\documentclass[10pt,twocolumn,letterpaper]{article}

\usepackage{iccv}      %

\usepackage{amssymb}%
\usepackage{pifont}%
\usepackage{makecell}
\usepackage{tabu, booktabs}
\usepackage{microtype}
\usepackage{colortbl}
\usepackage{xcolor}
\usepackage[pagebackref,breaklinks,colorlinks,citecolor=cvprblue]{hyperref}
\usepackage[accsupp]{axessibility}

\usepackage{algorithm}
\usepackage{algpseudocode}
\usepackage{graphicx} %
\usepackage{tikz}     %

\newcommand{\figref}[1]{Fig.~\ref{#1}}
\newcommand{\secref}[1]{Sec.~\ref{#1}}
\newcommand{\eqnref}[1]{Eq.~\eqref{#1}}
\newcommand{\tabref}[1]{Table~\ref{#1}}

\newcommand{\bC}{\mathbf{C}}

\newcommand{\bI}{\mathbf{I}}

\newcommand{\cmark}{\textcolor{green}{\checkmark}} %
\newcommand{\xmark}{\textcolor{red}{\ding{55}}} %

\definecolor{iccvblue}{rgb}{0.21,0.49,0.74}
\definecolor{cvprblue}{rgb}{0.21,0.49,0.74}

\def\paperID{6761} %
\def\confName{ICCV}
\def\confYear{2025}

\title{CL-Splats: Continual Learning of Gaussian Splatting with Local Optimization}
\author{Jan Ackermann$^{1,2}$~~~Jonas Kulhanek$^{1,3}$~~~Shengqu Cai$^{2}$~~~Haofei Xu$^1$\\
Marc Pollefeys$^1$~~~Gordon Wetzstein$^2$~~~Leonidas J. Guibas$^{2,4}$~~~Songyou Peng$^4$\\
\\[-0.3cm]
$\phantom{}^1$ETH Zürich~~~~~$\phantom{}^2$Stanford University~~~~~$\phantom{}^3$CTU Prague~~~~~$\phantom{}^4$Google DeepMind
\vspace{-1.2cm}
}
\newcommand{\methodname}{{{CL-Splats}}\xspace}

\newread\imgstream
\immediate\openin\imgstream=imagedata.in
\makeatletter
\def\new@kvginclip#1{}
\def\new@kvgintrim#1{}
\let\old@kvginclip\KV@Gin@clip
\let\old@kvgintrim\KV@Gin@trim
\let\oldincludegraphics\includegraphics
\providecommand{\includegraphics}{}
\renewcommand{\includegraphics}[2][]{%
  \immediate\read\imgstream to \src
  \immediate\read\imgstream to \removecrop
  \ifnum\removecrop=1
      \let\KV@Gin@clip\new@kvginclip
      \let\KV@Gin@trim\new@kvgintrim
  \fi
  \oldincludegraphics[#1]{\src}%
  \let\KV@Gin@clip\old@kvginclip
  \let\KV@Gin@trim\old@kvgintrim}
\makeatother

\begin{document}
\twocolumn[{
\renewcommand\twocolumn[1][]{#1}%
\maketitle
\begin{center}
    \centering
\includegraphics[width=\textwidth]{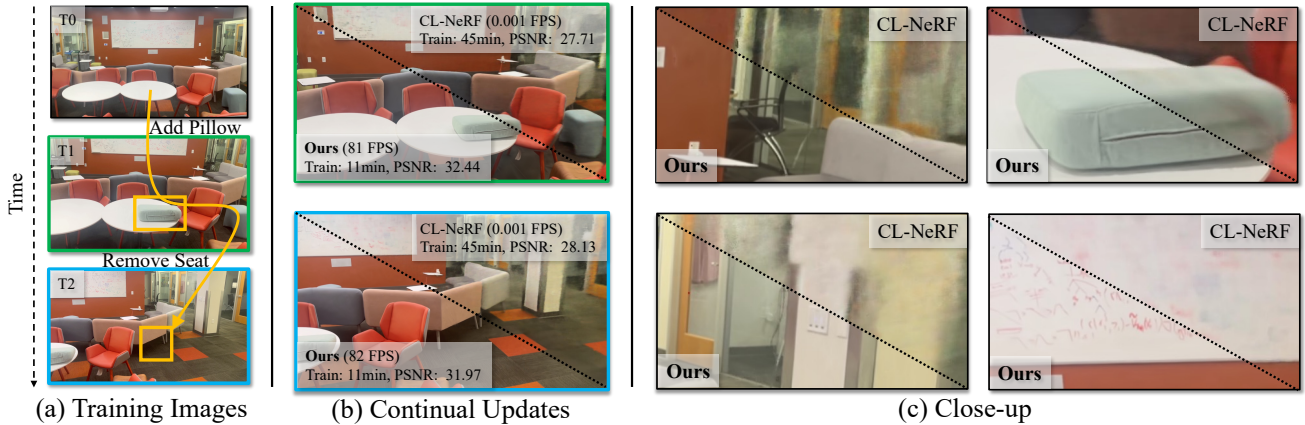}
    \vspace{-0.7cm}
    \captionof{figure}{
    We introduce \textbf{\methodname}, a simple yet effective approach for continual learning of Gaussian Splats. Our method efficiently updates scenes using a small, incremental set of views focused on changed regions. It achieves high-quality reconstructions while significantly outperforming existing continual learning methods for novel view synthesis in both reconstruction accuracy and optimization speed.}
    \label{fig:teaser}
\end{center}%
\vspace{0.1cm}
}]
\maketitle
\begin{abstract}
In dynamic 3D environments, accurately updating scene representations over time is crucial for applications in robotics, mixed reality, and embodied AI. As scenes evolve, efficient methods to incorporate changes are needed to maintain up-to-date, high-quality reconstructions without the computational overhead of re-optimizing the entire scene.
This paper introduces CL-Splats, which incrementally updates Gaussian splatting-based 3D representations from sparse scene captures.
CL-Splats integrates a robust change-detection module that segments updated and static components within the scene, enabling focused, local optimization that avoids unnecessary re-computation.
Moreover, CL-Splats supports storing and recovering previous scene states, facilitating temporal segmentation and new scene-analysis applications.
Our extensive experiments demonstrate that CL-Splats achieves efficient updates with improved reconstruction quality over the state-of-the-art. This establishes a robust foundation for future real-time adaptation in 3D scene reconstruction tasks.
We will release our source code and the synthetic and real-world datasets we created at \url{https://cl-splats.github.io/}\ .
\end{abstract}    
\section{Introduction}
\label{sec:intro}
\vspace{-0.3cm}
3D scene reconstruction is a fundamental challenge in computer vision and robotics~\cite{seitz2006comparison}, with applications across mixed reality, autonomous navigation, and embodied AI. Many tasks within these fields, such as object interaction or localization, benefit from a continually evolving understanding of the environment~\cite{wald2019rio}.
As scenes change over time, efficiently incorporating updates from new observations is essential for maintaining accurate and up-to-date reconstructions. These new updates can provide insights into object movements and transformations, enhancing our understanding of dynamic environments.

While real-time scene reconstruction has garnered substantial attention, techniques that address long-term scene evolution are less developed~\cite{sun2025nothing}.
Updating 3D representations from intermittent, sparse captures—rather than continuous video—introduces unique challenges that have to be addressed to improve dynamic scene understanding.
To this end, recent approaches have explored updating point clouds by matching and registering objects over time~\cite{zhu2024living}, but they lack photorealism and only reconstruct objects.
Other works introduce novel view synthesis techniques to update the reconstruction iteratively~\cite{cai2023clnerf, wu2024cl}, but they face issues with catastrophic forgetting and recovering of previous states, making them less versatile.
This highlights the need for a unified approach that effectively integrates update mechanisms while preserving the visual fidelity and efficiency of modern 3D scene representations such as Gaussian Splatting.

Our method, namely \methodname, combines the strengths of point-cloud-based and NeRF-based~\cite{mildenhall2021nerf} methods for continual scene updates, offering local and efficient optimization while achieving photorealistic reconstructions using the Gaussian Splatting~\cite{kerbl20233d} framework.
Compared to previous state-of-the-art methods and static-scene baselines, \methodname achieves more efficient updates and higher reconstruction quality.
Furthermore, our method enables segmenting and storing scene changes across time, supporting existing and emerging scene analysis applications.

Our key contributions are: 
\begin{itemize}
\item We introduce a general framework for continual and localized updates in Gaussian Splatting representations, enabling flexible and efficient scene reconstruction.
\item We present a novel local Gaussian optimization strategy which restricts the optimization of Gaussians in 3D space and enables efficient and local updates.
\item We empirically validate our approach on diverse real and synthetic scenes, demonstrating its effectiveness. Additionally, we show its ability to perform tasks such as batched updates and scene history recovery.
\item We contribute novel synthetic and real-world datasets designed explicitly for benchmarking dynamic scene reconstruction.
\end{itemize}
\begin{table}[t]
    \centering
    \resizebox{\linewidth}{!}{%
    \begin{tabular}{lcccc}
    \toprule
        Method & No Forgetting & History & Local & Time \\
        \midrule
        CLNeRF~\cite{cai2023clnerf} & \xmark & \xmark & \xmark & Slow \\
        CL-NeRF~\cite{wu2024cl} & \xmark & $\sim$ & \xmark & Slow \\ 
        \methodname \textbf{(Ours)} & \cmark & \cmark & \cmark & Fast \\
      \bottomrule
    \end{tabular}
    }
    \vspace{-0.25cm}
    \caption{\textbf{Overview of Related Methods.} Our explicit representation is the only that does not suffer from catastrophic forgetting while enabling an efficient and scalable history and optimizing locally, allowing new applications and the fastest optimization.}
    \label{tab:method_comp}
\end{table}
\vspace{-0.2cm}
\section{Related Work}
\label{sec:related}
\vspace{-0.2cm}
\begin{figure*}[!t]
    \centering
\includegraphics[width=\linewidth]{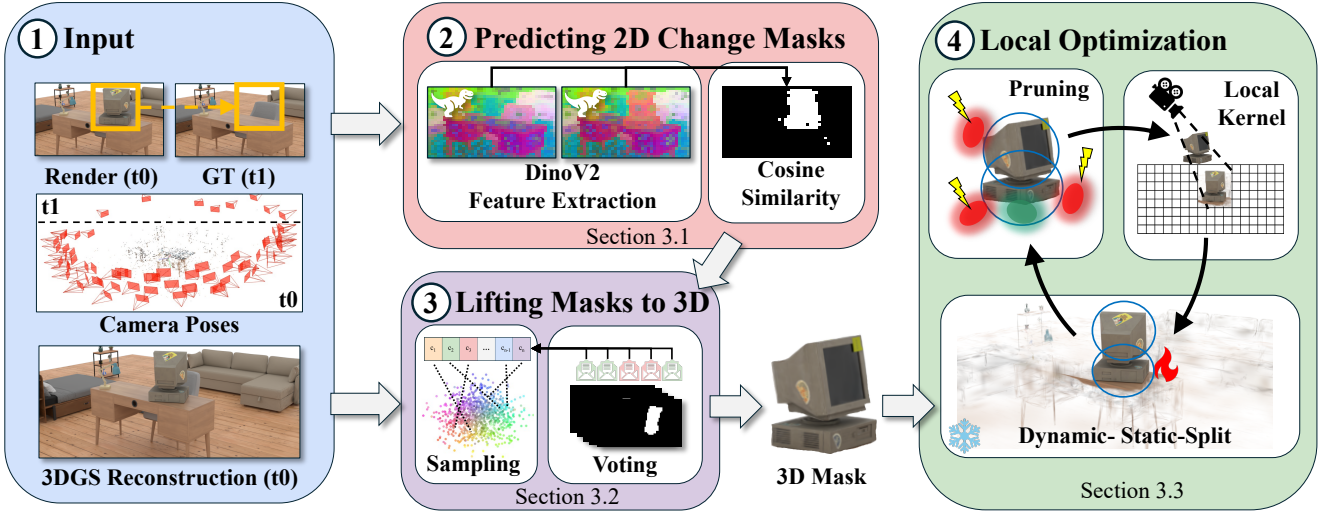}
    \vspace{-0.6cm}
    \caption{\textbf{Overview.} Starting from an existing reconstruction, we estimate the changed regions in 2D and then lift that information into 3D. After that we only optimize a local region around the changed parts. Our optimization is faster while obtaining the same gradients as 3DGS.}
    \label{fig:pipeline}
    \vspace{-0.4cm}
\end{figure*}

\paragraph{Continual Scene Reconstruction.}
Dynamic scene reconstruction methods aim to continuously update 3D representations, enabling systems to learn and adapt over time~\cite{li2024activesplat,cai2023clnerf, wu2024cl}. Significant research in this area has focused on photorealistic reconstructions from video sequences using point clouds and radiance fields.
Recent methods enhance scene representations incrementally: Zhou et al.~\cite{zhou2024drivinggaussian} introduce static Gaussians to explore new areas, Po et al.~\cite{po2023instant} refine scenes by resampling NeRF views, and Wang et al.~\cite{wang2024scarf} improve memory efficiency by decomposing scenes into compact NeRF components. However, these approaches focus on real-time view additions rather than handling long-term scene changes.
To address longer temporal differences, datasets and methods incorporating point cloud instance re-identification have been proposed~\cite{zhu2024living, wald2019rio, halber2019rescan, sun2023nothing}. 
Recent works CL-NeRF~\cite{wu2024cl} and CLNeRF~\cite{cai2023clnerf} enable photorealistic reconstruction for continual learning over extended periods and at discrete time steps.
Our approach builds upon these advancements, combining CL-NeRF’s photorealism with Gaussian Splatting’s adaptable representations to achieve high-quality rendering and efficient updates. In comparison to CLNeRF~\cite{cai2023clnerf}, our method only requires frames from the changed region, making it more efficient, and our method does not require knowing camera poses of unchanged regions, unlike CL-NeRF~\cite{wu2024cl}. Additionally, our localized and interpretable optimization enables efficient and accurate scene history recovery applications.
We summarize key differences between our method and prior work in \tabref{tab:method_comp}.

\vspace{-0.4cm}
\paragraph{3D Editing.}
Editing 3D scenes shares similarities with our work but differs fundamentally: editing is user-driven, while our method adapts to real-world scene changes based on observations.
Traditional tools like Maya~\cite{maya,blender} provide complete scene control, but many modern 3D representations, such as radiance fields, are implicit and not easily manipulable using conventional editing workflows. Consequently, recent research on editing radiance fields has focused on preserving static elements while selectively updating specific regions~\cite{chen2024gaussianeditor,yu2024cogs,ye2023gaussian,wang2024gaussianeditor,wang2024view,haque2023instruct,zhang2024stylizedgs,yuan2022nerf,song2023blending,dong2024vica,bao2023sine}.
While these methods effectively handle local edits, they often struggle with geometry modifications and maintaining high realism. In contrast, our approach leverages real-world scene updates to ensure photorealistic consistency, avoiding the inconsistencies introduced by 2D foundation models in synthetic editing tasks.
Additionally, recent techniques have explored object removal~\cite{wang2024learning, mirzaei2024reffusion, shen2024flashsplat}, manipulation~\cite{luo2024unsupervised}, and creation~\cite{chen2024gaussianeditor}. While these methods enable significant user-driven geometric changes, they remain limited to specific edit types. In contrast, our method is designed to accommodate general scene changes over time, allowing for continual, observation-driven updates without manual intervention.

\vspace{-0.4cm}
\paragraph{Scene Reconstruction from Sparse Views.}
NeRF~\cite{mildenhall2021nerf} and 3DGS~\cite{kerbl20233d} originally required densely sampled views~(around 100), which are impractical to obtain for many applications.
Recently, reconstruction and synthesis from sparse views~(e.g., 2 or 3) have gained attention~\cite{charatan2024pixelsplat,chen2021mvsnerf,chen2023explicit,niemeyer2022regnerf,szymanowicz2024flash3d,truong2023sparf,liu2024mvsgaussian,wu2024reconfusion,yu2022monosdf,xu2024murf,zhang2024gs,xu2024grm,yu2021pixelnerf,szymanowicz2024splatter,chen2024mvsplat,xu2024depthsplat,wang2023sparsenerf}.
These methods fall into two categories: per-scene optimization~\cite{chen2021mvsnerf,niemeyer2022regnerf,truong2023sparf,wu2024reconfusion,yu2022monosdf,chen2024mvsplat} and feed-forward inference~\cite{charatan2024pixelsplat,chen2023explicit,szymanowicz2024flash3d,xu2024murf,zhang2024gs,xu2024grm,yu2021pixelnerf,szymanowicz2024splatter,wang2023sparsenerf} .
The former uses robust regularization to enhance reconstruction. However, it is computationally intensive due to per-scene optimization. The latter enables efficient feed-forward reconstruction from sparse views by leveraging priors from large datasets but lacks generalization.
Unlike these methods, which hallucinate missing information due to extreme view sparsity, our approach leverages an existing reconstruction and assumes sparse views focused on local scene changes, significantly improving accuracy and photorealism.
\vspace{-0.2cm}
\section{Method}
\vspace{-0.2cm}
\label{sec:method}
Given an existing 3DGS reconstruction $\mathcal{G}^{t-1}$ at time $t-1$ and a new set of images $\bI^t = (I^t_0, I^t_1, \dots, I^t_n)$ capturing local scene changes at time $t$, our goal is to efficiently update $\mathcal{G}^{t-1}$ to reflect these changes. Crucially, our continual learning setting assumes that no past information beyond the current reconstruction $\mathcal{G}^{t-1}$ is available.
A naive approach would be to re-run 3DGS on $\bI^t$ to optimize a new reconstruction $\mathcal{G}^{t}$, assuming the new observations fully cover the scene. However, this method has major drawbacks.
Firstly, it fails to leverage the existing reconstruction $\mathcal{G}^{t-1}$, discarding unchanged information and making the optimization computationally inefficient.
Secondly, it requires re-capturing the entire scene, significantly reducing flexibility in dynamic settings.
Furthermore, blindly re-running 3DGS would erase parts of $\mathcal{G}^{t-1}$ in regions not sufficiently constrained by observations in $\bI^t$, leading to information loss.
To address these issues, our approach selectively updates only the necessary regions within $\mathcal{G}^{t-1}$, preserving unchanged areas and preventing unnecessary recomputation.

\vspace{-0.1cm}
\subsection{Detecting Changes in 2D} \label{sec:change_detection}
\vspace{-0.1cm}
Similar to existing 3D reconstruction methods, we first estimate camera poses $\bC^t = (c^t_0, c^t_1,...,c^t_n)$ for the images $\bI^{t}$, ensuring alignment with the existing reconstruction's coordinate system. We use COLMAP~\cite{schonberger2016structure} across our experiments to estimate the camera poses, following prior works~\cite{cai2023clnerf, wu2024cl}. We refer to supp.\ material regarding details about COLMAP setup on changing scenes.

Next, we identify changed regions in each $I^t_i \in \bI^t$ by comparing them to the existing reconstruction $\mathcal{G}^{t-1}$. We render a corresponding image $\hat{I}^{t-1}_i$ from $\mathcal{G}^{t-1}$ using the camera pose $c^t_i$ of $I^t_i$.
To detect changes, we leverage DINOv2's~\cite{oquab2023dinov2} feature extractor $\mathcal{E}$, to compute per-patch feature maps for both the rendered image $\hat{I}^{t-1}_i$ and the real image $I^{t}_i$:
\vspace{-0.1cm}
\begin{equation}
   \hat{\mathcal{F}}^{t-1}_i = \mathcal{E}(\hat{I}^{t-1}_i),\ \mathcal{F}^{t}_i = \mathcal{E}(I^{t}_i)
   \vspace{-0.1cm}
\end{equation}
Using these feature maps, we compute a 2D binary change mask $\mathcal{M}^{t}_i$ by thresholding the cosine similarity between corresponding feature vectors at each pixel:
\vspace{-0.1cm}
\begin{equation}
\mathcal{M}^{t}_i = \begin{cases}
1, & \text{if } \cos\left( \hat{\mathcal{F}}^{t-1}_i,\, \mathcal{F}^{t}_i \right) \leq \tau_1 \\
0, & \text{otherwise}
\label{eq:2d_mask}
\end{cases}
\vspace{-0.2cm}
\end{equation}

Finally, we resize the binary masks to match the original image resolution and dilate them by applying a sliding quadratic kernel over the entire frame. This step fills small noisy holes and expands the detected change regions, ensuring a more robust coverage of modified areas.

\vspace{-0.1cm}
\subsection{Lifting Changes to 3D}
\label{sec:identify_change_3d}
\vspace{-0.1cm}
While the change masks $\mathcal{M}^{t}_{i}$ from~\eqref{eq:2d_mask} capture local changes in image space, we observe that directly optimizing the 3D representation $\mathcal{G}^{t-1}$ using a photometric loss restricted to these 2D masks leads to poor results. This occurs because $\mathcal{M}^{t}_{i}$ lacks 3D spatial awareness, causing unintended updates to Gaussians that do not belong to the changed regions.
\vspace{-0.4cm}
\paragraph{Mask Majority Voting.}
To address this, we ensure that only Gaussians in the actual changed regions are optimized, preserving the rest of the scene. We achieve this through a majority voting mechanism that identifies which 3D Gaussians correspond to the detected changes.
Specifically, we project the Gaussians from $\mathcal{G}^{t-1}$ onto each 2D mask $\mathcal{M}^{t}_{i}$ and count how many times each Gaussian falls inside a mask across all views. Gaussians that appear within the masks in more than $K$ viewpoints are classified as part of the changed region. We denote these identified Gaussians as $\mathcal{O}^{t}$, which form the 3D change mask.
\vspace{-0.4cm}
\paragraph{Gaussian Sampling for New Objects.} While the above method works well for object removals, it fails when new objects appear, as there may be no existing 3D Gaussians in the affected region. To address this, we introduce a sampling strategy (detailed in Algorithm~\autoref{alg:new_points}) to ensure sufficient Gaussians in the changed regions. The subroutines are described in the supplementary material.
After sampling, we initialize all other 3DGS parameters following the standard SfM-based initialization. This guarantees that the method can properly model newly appearing objects. Importantly, we always sample these new points independently of the operation, maintaining flexibility in handling various types of scene changes.

\begin{algorithm}[t]
\caption{New Point Sampling}\label{alg:new_points}
\begin{algorithmic}[1]
    \State \textbf{Input:} $\mathcal{O}^{t}$, $\mathcal{G}^{t-1}$, $(\mathcal{M}^t_i)_{i\leq N}$, $n$
    \State \textbf{Output:} pointcloud $\mathcal{P}^t$
    \Function{SamplePoints}{$\mathcal{O}^{t}$, $\mathcal{G}^{t-1}$, $(\mathcal{M}^t_i)_{i\leq N}$,$n$}
        \If{$|\mathcal{O}^t| \geq n$}
            \State \Return $\mathcal{O}^t$
        \ElsIf{$|\mathcal{O}^t| = 0$}
            \State $\mathcal{S}=$ \Call{RandomSample}{$\mathcal{G}^{t-1}$, $(\mathcal{M}^t_i)_{i\leq N}$,
            $n$}
            \State \Return \Call{SamplePoints}{$\mathcal{S}$, $\mathcal{G}^{t-1}$, $(\mathcal{M}^t_i)_{i\leq N}$,$n$}
        \Else
            \State $\mathcal{S}=$ \Call{SampleRegion}{$\mathcal{O}^t$, $(\mathcal{M}^t_i)_{i\leq N}$, $n$}
            \State $\mathcal{O} = \mathcal{O}^t \cup \mathcal{S}$
            \State \Return \Call{SamplePoints}{$\mathcal{O}$, $\mathcal{G}^{t-1}$, $(\mathcal{M}^t_i)_{i\leq N}$,$n$}
        \EndIf
    \EndFunction
\end{algorithmic}
\end{algorithm}

\subsection{Local Optimization in Changed 3D Regions} \label{sec:localized_gaussian_optimization}
The final step involves optimizing $\mathcal{O}^t$ within the detected changed regions.
A straightforward approach would be to apply standard 3DGS optimization to all Gaussians in the scene. However, this leads to
(a) unnecessary computation over unchanged areas, making optimization inefficient,
and (b) sub-optimal results, as optimizing all Gaussians does not respect the localized nature of scene changes.
To address these issues, we propose a locally constrained and exact optimization scheme that ensures updates remain spatially accurate while eliminating boundary artifacts. The following section details our method.

\begin{figure}[!t]
    \centering
    \includegraphics[width=\linewidth]{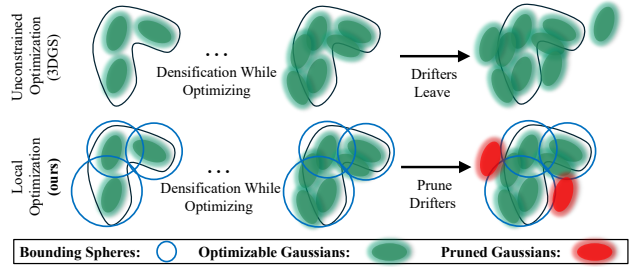}
    \vspace{-0.6cm}
    \caption{\textbf{Sphere Pruning.} We first obtain a dynamic rendering mask. Then we use this to efficiently compute the gradients. Finally, we prune gaussians that leave the 3D region of the change.}
    \label{fig:local_optimization}
\end{figure}

\paragraph{Sphere Pruning.} While determining which Gaussians project into a 2D mask is straightforward, enforcing spatial constraints in 3D presents two key challenges: (a) defining the 3D optimization region, and
(b) ensuring that Gaussians remain within this region throughout optimization.
To efficiently bound the optimization region, we represent it with a set of geometric primitives that enable sparse representation while allowing for fast membership checks. We choose spheres as our bounding primitives due to their:
1) ease of fitting,
2) suitability for Euclidean distance-based clustering algorithms,
and 3) efficiency to check whether a Gaussian is inside the region.
To obtain the spheres, we first identify clusters with HDBSCAN~\cite{mcinnes2017hdbscan}, perform outlier detection, then fit one sphere around each cleaned-up cluster. 
Notably, a single changing object may contain multiple clusters, requiring multiple spheres for accurate coverage. As shown in \figref{fig:local_optimization}, we enforce spatial constrains by pruning Gaussians once their centers move outside the union of bounding spheres, ensuring that optimization remains strictly localized.

\vspace{-0.4cm}

\paragraph{Local Optimization Kernel.} The achieved spatial locality significantly accelerates optimization by performing computations only on relevant Gaussians and tiles. We design an efficient optimization algorithm that computes the same gradients as standard 3DGS for all Gaussians in $\mathcal{O}^t$, while minimizing computational costs. Below, we provide a brief overview, with further details available in the supplementary material. 
At each optimization step, we:

1) Dynamically project the Gaussians $\mathcal{O}^t$ into the image plane, generating a 2D render mask.
2) Render and compute gradients exclusively for pixels within this masked region, avoiding unnecessary computations.
3) During backpropagation, update only the Gaussians contributing to the masked pixels, ensuring precise and efficient optimization.

Crucially, this localized approach preserves the exact gradients that $\mathcal{O}^t$ would receive under a full-scene optimization step, maintaining update quality while significantly reducing computational overhead.
\section{Experiments}
\label{sec:experiments}
\begin{table*}[t]
    \centering
    \begin{tabular}{lcccccccc}
    \toprule
        & \multicolumn{4}{c}{\textbf{Synthetic}} & \multicolumn{4}{c}{\textbf{Real World}}\\
         \cmidrule(lr){2-5} \cmidrule(lr){6-9}
         Method &  PSNR$\uparrow$ & LPIPS$\downarrow$ & SSIM$\uparrow$ & FPS$\uparrow$ & PSNR$\uparrow$ & LPIPS$\downarrow$ & SSIM$\uparrow$ & FPS$\uparrow$ \\\midrule
    3DGS~\cite{kerbl20233d} & 21.993 & 0.189 & 0.838 & 221 & 11.764 & 0.376 & 0.399 &  125\\
    3DGS+M & 15.127 & 0.303 & 0.737 & \textbf{254} & 8.585 & 0.461 & 0.271 & \textbf{151}\\
    GaussianEditor~\cite{chen2024gaussianeditor} & 19.801 & 0.197 & 0.871 & \underline{227} & 24.133 & \underline{0.143} & \underline{0.867} & \underline{137}\\
    CLNeRF~\cite{cai2023clnerf} & 26.758 & 0.322 & 0.738 & $<$1 & \underline{24.541} & 0.373 & 0.658 & $<$1 \\
    CL-NeRF~\cite{wu2024cl} & \underline{30.063} & \underline{0.058} & \underline{0.939} & $<$1 & 23.268 & 0.290 & 0.725 & $<$1\\ 
      \methodname \textbf{(ours)}   &  \textbf{40.125} & \textbf{0.015}& \textbf{0.985} & 223 & \textbf{28.249} & \textbf{0.065} & \textbf{0.930} & 135 \\
      \bottomrule
    \end{tabular}
    \vspace{-0.1cm}
    \caption{\textbf{Novel View Synthesis Results on Our \textit{CL-Splats} Dataset.} The best metric is marked in bold and the second best is underlined.
     The resolution for all images are $960\times 540$.
    }
    \label{tab:main_table}
    \vspace{-0.3cm}
\end{table*}

\begin{figure*}[ht]
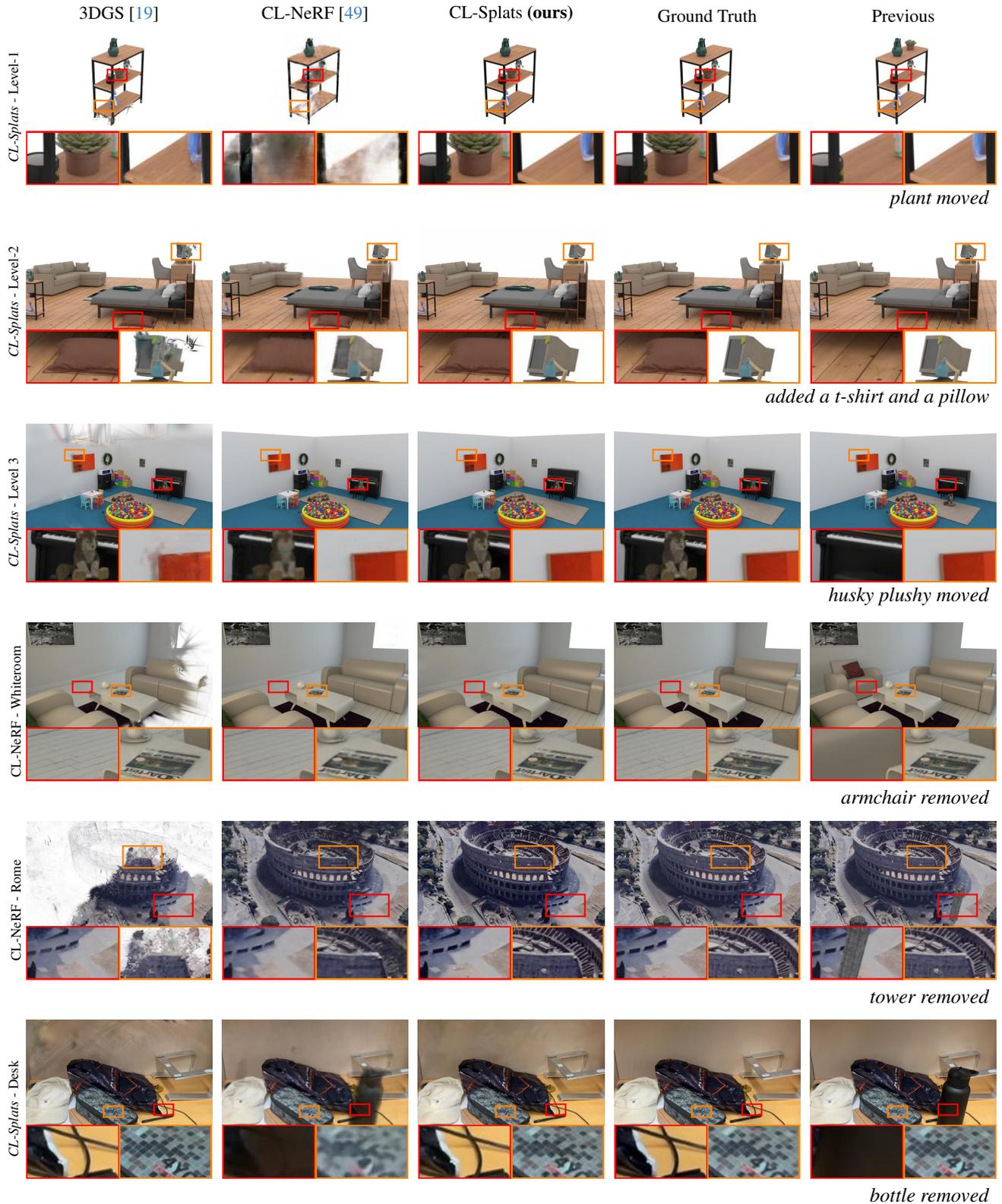

    \centering
    \pgfmathsetmacro{\imageWidth}{960}
\pgfmathsetmacro{\imageHeight}{540}
\pgfmathsetmacro{\scaleFactorX}{0.24 * \textwidth / \imageWidth}
\pgfmathsetmacro{\scaleFactorY}{0.24 * \textwidth / \imageHeight}
\pgfmathsetmacro{\cropWidth}{100 / \imageWidth}
\pgfmathsetmacro{\cropx}{420 / \imageWidth}
\pgfmathsetmacro{\cropy}{260 / \imageHeight}
\pgfmathsetmacro{\scropx}{350 / \imageWidth}
\pgfmathsetmacro{\scropy}{100/ \imageHeight}

\begin{tikzpicture}
    \newcommand\cscale{0.1865}
    \node[] (image) at (0,0) {};
    \newcommand\expandImage{
        \begin{scope}[shift={(image.south west)}]
        \begin{scope}[shift={(image.south west)},x={(image.south east)},y={(image.north west)}]
            \pgfmathsetmacro{\xtwo}{\cropx + \cropWidth}
            \pgfmathsetmacro{\ytwo}{\cropy + \cropWidth}
            \draw[red, thick] (\cropx,\cropy) rectangle (\xtwo,\ytwo);
            \pgfmathsetmacro{\xtwo}{\scropx + \cropWidth}
            \pgfmathsetmacro{\ytwo}{\scropy + \cropWidth}
            \draw[orange, thick] (\scropx,\scropy) rectangle (\xtwo,\ytwo);
        \end{scope}
        \pgfmathsetmacro{\abscropx}{\cropx * \imageWidth}
        \pgfmathsetmacro{\abscropy}{\cropy * \imageHeight}
        \pgfmathsetmacro{\cropt}{(1 - (\cropx + \cropWidth)) * \imageWidth}%
        \pgfmathsetmacro{\cropl}{(1 - (\cropy + \cropWidth)) * \imageHeight}
    
        \node[anchor=north west,inner sep=0,draw=red,line width=2pt] (bottom_image) at (image.south west) {
         \includegraphics[width=0.0895\textwidth,trim={{\abscropx} {\abscropy} {\cropt} {\cropl}},clip]{\imagePath}};
         
        \pgfmathsetmacro{\abscropx}{\scropx * \imageWidth}
        \pgfmathsetmacro{\abscropy}{\scropy * \imageHeight}
        \pgfmathsetmacro{\cropt}{(1 - (\scropx + \cropWidth)) * \imageWidth}%
        \pgfmathsetmacro{\cropl}{(1 - (\scropy + \cropWidth)) * \imageHeight}
        \node[anchor=north east,inner sep=0,draw=orange,line width=2pt] (bottom_r_image) at (image.south east) {
         \includegraphics[width=0.0895\textwidth,trim={{\abscropx} {\abscropy} {\cropt} {\cropl}},clip]{\imagePath}
        };
        \end{scope}
    }

    \def\imagePath{images/main/level-1-gs}
    \node[anchor=north west,inner sep=0,alias=image,line width=0pt] (image1) at (0,0) {\includegraphics[width=\cscale\textwidth]{\imagePath}};
    \node[anchor=south,yshift=-15pt,xshift=1pt,rotate=90] at (image1.west) 
    {{\scriptsize{} \textit{CL-Splats} - Level-1}};
    \node[anchor=south, inner sep=2pt] at (image.north) {\small 3DGS~\cite{kerbl20233d}};
    \expandImage
    \node[anchor=south west] (row-marker) at (bottom_image.south west) {};
    
    \def\imagePath{images/main/cl-nerf-1}
    \node[anchor=south west,inner sep=0,alias=image,line width=0pt,xshift=0.01\textwidth] (image2) at (image.south east) {\includegraphics[width=\cscale\textwidth]{\imagePath}};
    \node[anchor=south, inner sep=2pt] at (image.north) {\small CL-NeRF~\cite{wu2024cl}};
    \expandImage

    \def\imagePath{images/main/ours_1_move.jpg}
    \node[anchor=south west,inner sep=0,alias=image,line width=0pt,xshift=0.01\textwidth] (image2) at (image.south east) {\includegraphics[width=\cscale\textwidth]{\imagePath}};
    \node[anchor=south, inner sep=2pt] at (image.north) {\small \methodname \textbf{(ours)}};
    \expandImage
    
    \def\imagePath{images/main/gt_1}
    \node[anchor=south west,inner sep=0,alias=image,line width=0pt,xshift=0.01\textwidth] (image3) at (image.south east) {\includegraphics[width=\cscale\textwidth]{\imagePath}};
    \node[anchor=south, inner sep=2pt] at (image.north) {\small Ground Truth};
    \expandImage

    \def\imagePath{images/main/previous-level-1}
    \node[anchor=south west,inner sep=0,alias=image,line width=0pt,xshift=0.01\textwidth] (image4) at (image.south east) {\includegraphics[width=\cscale\textwidth]{\imagePath}};
    \node[anchor=south, inner sep=2pt] at (image.north) {\small Previous};
    \expandImage
    
    \node[anchor=south east,yshift=-15] at (bottom_r_image.south east) {\textit{plant moved}};

\newcommand\linegap{-20}
    
\pgfmathsetmacro{\imageWidth}{960}
\pgfmathsetmacro{\imageHeight}{540}
\pgfmathsetmacro{\scaleFactorX}{0.24 * \textwidth / \imageWidth}
\pgfmathsetmacro{\scaleFactorY}{0.24 * \textwidth / \imageHeight}
\pgfmathsetmacro{\cropWidth}{150 / \imageWidth}
\pgfmathsetmacro{\cropx}{450 / \imageWidth}
\pgfmathsetmacro{\cropy}{5 / \imageHeight}
\pgfmathsetmacro{\scropx}{750 / \imageWidth}
\pgfmathsetmacro{\scropy}{360 / \imageHeight}

\def\imagePath{images/main/level-2-gs}
\node[anchor=north west,inner sep=0,alias=image,line width=0pt,yshift=\linegap] (image1) at (row-marker.south west) {\includegraphics[width=\cscale\textwidth]{\imagePath}};
\expandImage
\node[anchor=south west] (row-marker) at (bottom_image.south west) {};
\node[anchor=south,yshift=-13pt,xshift=1pt,rotate=90] at (image1.west) 
{{\scriptsize{} \textit{CL-Splats} - Level-2}};

\def\imagePath{images/main/cl-nerf-2-multi-9}
\node[anchor=south west,inner sep=0,alias=image,line width=0pt,xshift=0.01\textwidth] (image2) at (image.south east) {\includegraphics[width=\cscale\textwidth]{\imagePath}};
\expandImage

\def\imagePath{images/main/ours_2_multi_9}
\node[anchor=south west,inner sep=0,alias=image,line width=0pt,xshift=0.01\textwidth] (image2) at (image.south east) {\includegraphics[width=\cscale\textwidth]{\imagePath}};
\expandImage

\def\imagePath{images/main/gt_2_multi_9}
\node[anchor=south west,inner sep=0,alias=image,line width=0pt,xshift=0.01\textwidth] (image3) at (image.south east) {\includegraphics[width=\cscale\textwidth]{\imagePath}};
\expandImage

\def\imagePath{images/main/previous-level-2}
\node[anchor=south west,inner sep=0,alias=image,line width=0pt,xshift=0.01\textwidth] (image4) at (image.south east) {\includegraphics[width=\cscale\textwidth]{\imagePath}};
\expandImage

    \node[anchor=south east,yshift=-15] at (bottom_r_image.south east) {\textit{added a t-shirt and a pillow}};

\pgfmathsetmacro{\imageWidth}{960}
\pgfmathsetmacro{\imageHeight}{540}
\pgfmathsetmacro{\scaleFactorX}{0.24 * \textwidth / \imageWidth}
\pgfmathsetmacro{\scaleFactorY}{0.24 * \textwidth / \imageHeight}
\pgfmathsetmacro{\cropWidth}{100 / \imageWidth}
\pgfmathsetmacro{\cropx}{650 / \imageWidth}
\pgfmathsetmacro{\cropy}{200 / \imageHeight}
\pgfmathsetmacro{\scropx}{200 / \imageWidth}
\pgfmathsetmacro{\scropy}{350 / \imageHeight}

\def\imagePath{images/main/level-3-gs.jpg}
\node[anchor=north west,inner sep=0,alias=image,line width=0pt,yshift=\linegap] (image1) at (row-marker.south west) {\includegraphics[width=\cscale\textwidth]{\imagePath}};
\expandImage
\node[anchor=south west] (row-marker) at (bottom_image.south west) {};
\node[anchor=south,yshift=-19pt,xshift=1pt,rotate=90] at (image1.west) 
{{\scriptsize{} \textit{CL-Splats} - Level 3}};

\def\imagePath{images/main/cl-nerf-3-move.jpg}
\node[anchor=south west,inner sep=0,alias=image,line width=0pt,xshift=0.01\textwidth] (image2) at (image.south east) {\includegraphics[width=\cscale\textwidth]{\imagePath}};
\expandImage

\def\imagePath{images/main/ours_3_move.jpg}
\node[anchor=south west,inner sep=0,alias=image,line width=0pt,xshift=0.01\textwidth] (image2) at (image.south east) {\includegraphics[width=\cscale\textwidth]{\imagePath}};
\expandImage

\def\imagePath{images/main/gt_3_move}
\node[anchor=south west,inner sep=0,alias=image,line width=0pt,xshift=0.01\textwidth] (image3) at (image.south east) {\includegraphics[width=\cscale\textwidth]{\imagePath}};
\expandImage

\def\imagePath{images/main/previous-level-3}
\node[anchor=south west,inner sep=0,alias=image,line width=0pt,xshift=0.01\textwidth] (image4) at (image.south east) {\includegraphics[width=\cscale\textwidth]{\imagePath}};
\expandImage

    \node[anchor=south east,yshift=-15] at (bottom_r_image.south east) {\textit{husky plushy moved}};

\pgfmathsetmacro{\imageWidth}{960}
\pgfmathsetmacro{\imageHeight}{540}
\pgfmathsetmacro{\scaleFactorX}{0.24 * \textwidth / \imageWidth}
\pgfmathsetmacro{\scaleFactorY}{0.24 * \textwidth / \imageHeight}
\pgfmathsetmacro{\cropWidth}{100 / \imageWidth}
\pgfmathsetmacro{\cropx}{240 / \imageWidth}
\pgfmathsetmacro{\cropy}{180 / \imageHeight}
\pgfmathsetmacro{\scropx}{440 / \imageWidth}
\pgfmathsetmacro{\scropy}{160 / \imageHeight}

\def\imagePath{images/main/gs-whiteroom.jpg}
\node[anchor=north west,inner sep=0,alias=image,line width=0pt,yshift=\linegap] (image1) at (row-marker.south west) {\includegraphics[width=\cscale\textwidth]{\imagePath}};
\expandImage
\node[anchor=south west] (row-marker) at (bottom_image.south west) {};
\node[anchor=south,yshift=-19pt,xshift=1pt,rotate=90] at (image1.west) 
{{\scriptsize{} CL-NeRF - Whiteroom}};

\def\imagePath{images/main/cl-nerf-whiteroom.jpg}
\node[anchor=south west,inner sep=0,alias=image,line width=0pt,xshift=0.01\textwidth] (image2) at (image.south east) {\includegraphics[width=\cscale\textwidth]{\imagePath}};
\expandImage

\def\imagePath{images/main/ours-whiteroom.jpg}
\node[anchor=south west,inner sep=0,alias=image,line width=0pt,xshift=0.01\textwidth] (image2) at (image.south east) {\includegraphics[width=\cscale\textwidth]{\imagePath}};
\expandImage

\def\imagePath{images/main/gt-whiteroom.jpg}
\node[anchor=south west,inner sep=0,alias=image,line width=0pt,xshift=0.01\textwidth] (image3) at (image.south east) {\includegraphics[width=\cscale\textwidth]{\imagePath}};
\expandImage

\def\imagePath{images/main/whiteroom-before.jpg}
\node[anchor=south west,inner sep=0,alias=image,line width=0pt,xshift=0.01\textwidth] (image4) at (image.south east) {\includegraphics[width=\cscale\textwidth]{\imagePath}};
\expandImage

    \node[anchor=south east,yshift=-15] at (bottom_r_image.south east) {\textit{armchair removed}};

\pgfmathsetmacro{\imageWidth}{960}
\pgfmathsetmacro{\imageHeight}{540}
\pgfmathsetmacro{\scaleFactorX}{0.24 * \textwidth / \imageWidth}
\pgfmathsetmacro{\scaleFactorY}{0.24 * \textwidth / \imageHeight}
\pgfmathsetmacro{\cropWidth}{200 / \imageWidth}
\pgfmathsetmacro{\cropx}{660 / \imageWidth}
\pgfmathsetmacro{\cropy}{50 / \imageHeight}
\pgfmathsetmacro{\scropx}{500 / \imageWidth}
\pgfmathsetmacro{\scropy}{300 / \imageHeight}

\def\imagePath{images/main/3dgs-rome.jpg}
\node[anchor=north west,inner sep=0,alias=image,line width=0pt,yshift=\linegap] (image1) at (row-marker.south west) {\includegraphics[width=\cscale\textwidth]{\imagePath}};
\expandImage
\node[anchor=south west] (row-marker) at (bottom_image.south west) {};

\node[anchor=south west] (row-marker) at (bottom_image.south west) {};
\node[anchor=south,yshift=-19pt,xshift=1pt,rotate=90] at (image1.west) 
{{\scriptsize{} CL-NeRF - Rome}};

\def\imagePath{images/main/cl-nerf-rome.jpg}
\node[anchor=south west,inner sep=0,alias=image,line width=0pt,xshift=0.01\textwidth] (image2) at (image.south east) {\includegraphics[width=\cscale\textwidth]{\imagePath}};
\expandImage

\def\imagePath{images/main/ours-rome.jpg}
\node[anchor=south west,inner sep=0,alias=image,line width=0pt,xshift=0.01\textwidth] (image2) at (image.south east) {\includegraphics[width=\cscale\textwidth]{\imagePath}};
\expandImage

\def\imagePath{images/main/gt-rome.jpg}
\node[anchor=south west,inner sep=0,alias=image,line width=0pt,xshift=0.01\textwidth] (image3) at (image.south east) {\includegraphics[width=\cscale\textwidth]{\imagePath}};
\expandImage

\def\imagePath{images/main/rome-before.jpg}
\node[anchor=south west,inner sep=0,alias=image,line width=0pt,xshift=0.01\textwidth] (image4) at (image.south east) {\includegraphics[width=\cscale\textwidth]{\imagePath}};
\expandImage

    \node[anchor=south east,yshift=-15] at (bottom_r_image.south east) {\textit{tower removed}};

\pgfmathsetmacro{\imageWidth}{960}
\pgfmathsetmacro{\imageHeight}{540}
\pgfmathsetmacro{\scaleFactorX}{0.24 * \textwidth / \imageWidth}
\pgfmathsetmacro{\scaleFactorY}{0.24 * \textwidth / \imageHeight}
\pgfmathsetmacro{\cropWidth}{100 / \imageWidth}
\pgfmathsetmacro{\cropx}{660 / \imageWidth}
\pgfmathsetmacro{\cropy}{50 / \imageHeight}
\pgfmathsetmacro{\scropx}{400 / \imageWidth}
\pgfmathsetmacro{\scropy}{40 / \imageHeight}

\def\imagePath{images/main/3dgs-desk.jpg}
\node[anchor=north west,inner sep=0,alias=image,line width=0pt,yshift=\linegap] (image1) at (row-marker.south west) {\includegraphics[width=\cscale\textwidth]{\imagePath}};
\expandImage
\node[anchor=south west] (row-marker) at (bottom_image.south west) {};

\node[anchor=south west] (row-marker) at (bottom_image.south west) {};
\node[anchor=south,yshift=-19pt,xshift=1pt,rotate=90] at (image1.west) 
{{\scriptsize{} \textit{CL-Splats} - Desk}};

\def\imagePath{images/main/cl-nerf-desk.jpg}
\node[anchor=south west,inner sep=0,alias=image,line width=0pt,xshift=0.01\textwidth] (image2) at (image.south east) {\includegraphics[width=\cscale\textwidth]{\imagePath}};
\expandImage

\def\imagePath{images/main/ours-desk.jpg}
\node[anchor=south west,inner sep=0,alias=image,line width=0pt,xshift=0.01\textwidth] (image2) at (image.south east) {\includegraphics[width=\cscale\textwidth]{\imagePath}};
\expandImage

\def\imagePath{images/main/gt-desk.jpg}
\node[anchor=south west,inner sep=0,alias=image,line width=0pt,xshift=0.01\textwidth] (image3) at (image.south east) {\includegraphics[width=\cscale\textwidth]{\imagePath}};
\expandImage

\def\imagePath{images/main/before-desk.jpg}
\node[anchor=south west,inner sep=0,alias=image,line width=0pt,xshift=0.01\textwidth] (image4) at (image.south east) {\includegraphics[width=\cscale\textwidth]{\imagePath}};
\expandImage

    \node[anchor=south east,yshift=-15] at (bottom_r_image.south east) {\textit{bottle removed}};

\end{tikzpicture}
    \vspace{-1.5em}
    \caption{\textbf{Qualitative Comparison on All Datasets.} This figure shows renders of our method compared to 3DGS and CL-NeRF. The views are from the test-trajectory but here we show an angle that contains the change. The red crop always shows the area that has changed while the orange crop contains an unchanged region.
    We can see that our method faithfully produces high quality reconstructions while 3DGS destroys unconstrained regions and CL-NeRF gives less detailed reconstructions.
    }
    \label{fig:main_image}
    \vspace{-3.5em}
\end{figure*}

\paragraph{Datasets.}
We use three benchmarks across our evaluation.
Firstly, we evaluate on the synthetic data from CL-NeRF~\cite{wu2024cl}, with scenes originally from Mitsuba~\cite{nimier2019mitsuba}.
We adapt their data and use Mip-NeRF 360~\cite{barron2022mip}-style trajectories for the initial reconstruction.
We also build our \emph{CL-Splats} dataset containing two benchmarks with synthetic and real-world scenes.
The synthetic scenes contain three complexity levels, ranging from a simple scene with multiple objects through a one-room setting to an apartment containing four rooms. Each scene covers four scenarios: adding an object, removing an object, moving an object, and combinations.
We capture five diverse indoor and outdoor scenes, incorporating real-world changes such as object additions, removals, and rearrangements. The dataset includes scenarios with multiple concurrent object manipulations, providing a rigorous test of our method’s adaptability.

\paragraph{Baselines.}
We compare our approach with 3DGS~\cite{kerbl20233d}, CL-NeRF~\cite{wu2024cl}, CLNeRF~\cite{cai2023clnerf} and GaussianEditor~\cite{chen2024gaussianeditor} with the same 2D masks from~\secref{sec:change_detection}. 
In addition, we create a customized baseline called 3DGS+M, where we only apply the photometric loss within the 2D masks to optimize 3DGS.

\subsection{Evaluation}
\paragraph{Evaluation on \textit{CL-Splats} Dataset.}
As shown in~\tabref{tab:main_table}, our \methodname outperforms all baselines by a large margin in synthetic and real-world scenes while maintaining real-time rendering capabilities.
Although 3DGS+M uses the same 2D masks as our method, it performs significantly worse than the vanilla 3DGS.
This observation supports our earlier discussion in~\secref{sec:identify_change_3d} that locally optimizing the scene representation with photometric loss only within 2D masks could lead to undesirable artifacts.
In the qualitative comparison presented in~\figref{fig:main_image}, we observe apparent artifacts in the renderings produced by vanilla 3DGS, especially when the scene complexity increases, as shown in the last three rows. This happens particularly in less observed or unobserved areas in the sparse input views.
CL-NeRF~\cite{cai2023clnerf} preserves the unchanged regions, but its rendered views tend to be over-smoothed, as shown in the toy husky in the synthetic scene and the keyboard in the real-world scene. In contrast, our method captures significantly more details across all scenes.
We additionally perform dense captures of each synthetic scene after changes and run 3DGS for comparison. The average PSNR obtained is around 42 dB, an upper bound for reconstruction quality.
This further demonstrates the effectiveness of our \methodname, which achieves 40.1 dB, while the second-best method CL-NeRF~\cite{cai2023clnerf} only achieves 30.1 dB.
We also observe that object removal is usually the most straightforward task for both \methodname and CL-NeRF~\cite{cai2023clnerf} while handling multiple objects is the most challenging.

\paragraph{Evaluation on CL-NeRF Dataset.}
\begin{table}[!t]
    \centering
    \begin{tabular}{lccc}
    \toprule
         Method &  PSNR$\uparrow$ & LPIPS$\downarrow$ & SSIM$\uparrow$\\\midrule
    3DGS~\cite{kerbl20233d} & 11.072 & 0.356 & 0.537\\
    CL-NeRF~\cite{wu2024cl} & 27.302 & 0.177 & 0.829  \\ 
      \methodname \textbf{(ours)} & \textbf{29.984} & \textbf{0.156} & \textbf{0.839} \\
      \bottomrule
    \end{tabular}
    \caption{\textbf{Novel View Synthesis Results on the CL-NeRF Dataset.}
    Our method performs the best across all metrics.
    }
    \label{tab:reason_recov}
\end{table}
The CL-NeRF dataset poses several unique challenges: 1) its scenes contain complex surface reflectances, 2) two scenes in the dataset are primarily white and textureless in many regions~(e.g., the white room shown in~\figref{fig:main_image}), and 3) the scenes are enclosed by walls and ceilings that require a lot more 3D Gaussians for 3DGS-based methods to represent.
Due to these complexities, all methods generally exhibit lower performances on this dataset, as shown in \tabref{tab:reason_recov}. Nonetheless, our method still achieves superior quantitative performance despite these challenges.
Beyond indoor environments, the dataset also includes a large-scale outdoor scene depicting the Colosseum. As demonstrated in \figref{fig:main_image}, our method adapts to changes even in large-scale settings, further highlighting its robustness. It is worth noting that due to its underlying representation it would not easily scale to higher levels of details such as driving data.
For 3DGS-based approaches, we observe in both the Whiteroom and Rome scenes (\figref{fig:main_image}) that the reduced number of views destroys existing reconstruction, emphasizing the challenges posed by limited observations.

\subsection{Ablation Studies \& Analysis}
\paragraph{Ablation on Mask Quality.}
\begin{table}[t]
    \centering
    \begin{tabular}{lcc}
    \toprule
     Method & Recall$\uparrow$ & Precision$\uparrow$\\
    \midrule
    (Baseline) Color L2 & 0.761 & 0.281 \\
    (Baseline) SSIM & 0.745 & 0.332\\
    (a) $\mathcal{M}^{t}$ from~\eqnref{eq:2d_mask} & 0.961 & 0.370 \\
    (b) w/o. Dilation & 0.941 & 0.611\\
    (c) \textbf{Full (ours)}  & 0.942 & 0.609\\
      \bottomrule
      \vspace{0.1cm}
    \end{tabular}
    \setlength\tabcolsep{1pt} %
    \begin{tabular}{cc} %
        \makecell{\includegraphics[width=0.23\textwidth]{images/dilated.jpg}} & 
        \makecell{\includegraphics[width=0.23\textwidth]{images/non-dilated.jpg}} \\
        (a) & (b)\\
    
        \makecell{\includegraphics[width=0.23\textwidth]{images/ours-yay.jpg}} & 
        \makecell{\includegraphics[width=0.23\textwidth]{images/manual.jpg}} \\
        (c) & SAM Annotation
    \end{tabular}
    \vspace{-0.1cm}
    \caption{\textbf{Ablation on the Mask Quality.}
    (a) 2D masks from our 2D change detection module; (b) The 2D masks without dilation (in figure before reprojection); (c) Our reprojected masks from local optimization.
    }
    \label{tab:ablation_mask}
\end{table}
\label{sec:mask-ablation}
We investigate the impact of different kinds of masks: pixel-space baselines, the 2D masks from~\secref{sec:change_detection}, and the final masks from our reprojected changed 3D regions. 
We compute the precision and recall of those masks compared to manually annotated masks on our synthetic dataset on all three complexity levels.
As shown in \tabref{tab:ablation_mask}, our initial masks $\mathcal{M}^{t}$ from~\eqnref{eq:2d_mask} provide high recall but low precision.
Note that high recall is important for the optimization in the later stage because we need to guarantee that all changes are within the masks. This requirement makes the pixel-space baselines impractical.
After the local optimization in~\secref{sec:localized_gaussian_optimization}, even from these coarse and noisy 2D masks, we can obtain sharp 3D segmentation.
This way, we achieve good precision while maintaining high recall, as shown in~\tabref{tab:ablation_mask} (c).
A lower recall would lead to the failure of optimizing parts that would require to be changed.
A failure case is shown in the supplements.
We also investigate optimizing using the masks $\mathcal{M}^{t}$ without dilation applied.
As shown in~\tabref{tab:ablation_mask} (b), our approach is robust to the quality of the input masks.
In addition, we visualize the impact of our components on the estimated mask. We can observe that while (a) and (b) struggle with precision and recall, respectively, our mask (c) covers the changed region closely. Note that (c) predicts tiles since 3DGS gets optimized per tile, which leads to a slight mask expansion and a block appearance.

\begin{figure}
    \centering
    \includegraphics[width=\linewidth]{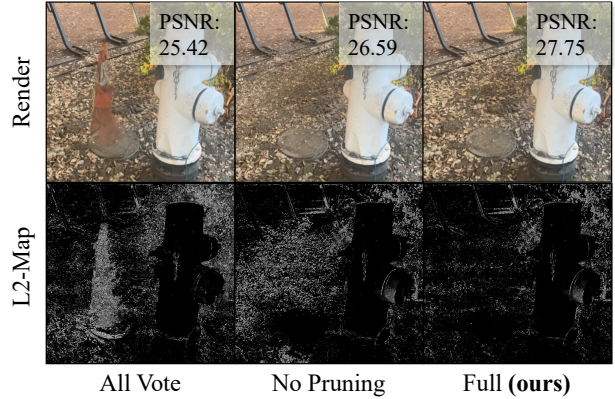}
    \vspace{-0.8cm}
    \caption{\textbf{Visual Comparison of Ablated Components.} The top row shows the optimized scene. The bottom shows the L2 difference multiplied by 50 for improved visibility.}
    \label{fig:qual-ablation}
\end{figure}

\begin{table}[t]
    \centering
    \setlength{\tabcolsep}{2pt}          %
    \resizebox{0.47\textwidth}{!}{%
    \begin{tabular}{lcccc}
        \toprule
        Method & PSNR$\uparrow$ & SSIM$\uparrow$ & LPIPS $\downarrow$ & Time \\
        \midrule 
        (a) No Frozen Background & 20.773 & 0.811 & 0.176 & 8 min \\
        (b) All Vote & 35.611 & 0.881 & 0.102 & 5min\\ 
        (c) w/o. Kernel & 40.812 & 0.978 & 0.018 & 8 min\\
        (d) Rectangle Primitive & 40.717 & 0.979 & 0.018 & 5 min \\
        (e) \textbf{Full (ours)} & 40.833 & 0.980 & 0.018 & 5 min \\
        \bottomrule
    \end{tabular}}
    \caption{\textbf{Ablation on Optimization.} The metrics were obtained on the test set of our \textit{CL-Splats} dataset -- Level 2.}
    \label{tab:local_optimization}
\end{table}
\paragraph{Ablation on Optimization.}
In \tabref{tab:local_optimization}, we conduct an ablation study to evaluate the impact of different core components in our local optimization framework (\secref{sec:localized_gaussian_optimization}).
The most critical factor affecting reconstruction quality is freezing all Gaussians outside the changed region. As seen in \tabref{tab:local_optimization} (a), without this mechanism, optimization degrades less-observed areas, highlighting its necessity.
We also compare our voting strategy to a stricter approach requiring Gaussians to be projected into all masks. As \figref{fig:qual-ablation} shows this leaves some points unoptimized, leading to lower reconstruction quality.
We find no significant difference in performance using oriented bounding boxes compared to our proposed spheres. However, membership checks for bounding boxes require three times more FLOPS, making spheres more efficient.
Additionally, we evaluate our local kernel design. It maximizes the efficiency of local optimization, reducing average optimization time by 60\% while maintaining reconstruction quality. A key factor in this speedup is dynamic reprojection (Step 1 in \secref{sec:localized_gaussian_optimization}), which restricts computations to tiles covered by changed Gaussians.

\paragraph{Fast Convergence.}
\begin{figure}[t!]
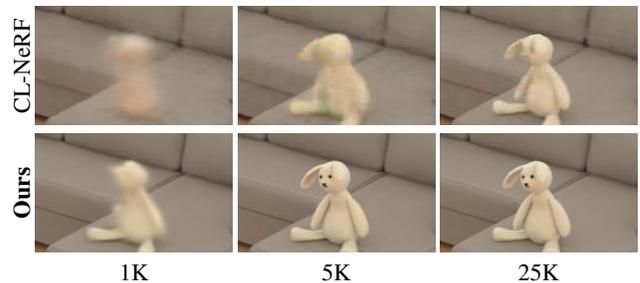

    \centering
    \setlength\tabcolsep{1pt} %
    \small
    \begin{tabular}{lccc} %
        \makecell{\rotatebox{90}{\small CL-NeRF}}&
        \makecell{\includegraphics[width=0.15\textwidth]{images/cl-1000.jpg}} & 
        \makecell{\includegraphics[width=0.15\textwidth]{images/cl-5000.jpg}} & 
        \makecell{\includegraphics[width=0.15\textwidth]{images/cl-25000.jpg}} \\
        \makecell{\rotatebox{90}{\small \textbf{Ours}}} &
        \makecell{\includegraphics[width=0.15\textwidth]{images/ours-1000.jpg}} & 
        \makecell{\includegraphics[width=0.15\textwidth]{images/ours-5000.jpg}} & 
        \makecell{\includegraphics[width=0.15\textwidth]{images/ours-25000.jpg}} \\
        & 1K & 5K & 25K\\
    \end{tabular}
    \vspace{-0.2cm}
    \caption{\textbf{Convergence Speed Comparison.} Our method can already capture much better details at 5K iterations than CL-NeRF at 25K iterations.}
    \label{fig:convergence}
\end{figure}
In \figref{fig:convergence}, we compare the training convergence of CL-NeRF~\cite{wu2024cl} and our \methodname on a synthetic scene from our \textit{CL-Splats} dataset.
Our \methodname achieves high-quality reconstruction in just 5K iterations, completing training in 40 seconds on an NVIDIA RTX Quadro 6000 GPU. In contrast, CL-NeRF~\cite{cai2023clnerf} requires 25K iterations and 50 minutes—75× slower—while still producing noticeably less detail in the final rendering.

\section{Applications}
\begin{figure}
    \centering
\includegraphics[width=\linewidth]{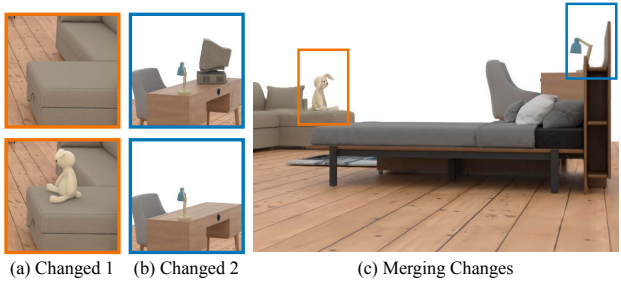}
\vspace{-0.6cm}
    \caption{\textbf{Merging Changes in Non-overlapping Areas.} 
    Here we show an example of our created synthetic scene.
    (a) A toy bunny is added onto the sofa. (b) A computer is removed from the desk. (c) Our method enables merging batches of changes in a single scene.
    }
    \label{fig:batched}
\end{figure}

\vspace{-0.1cm}
Our method uniquely localizes changes within the 3D Gaussian representation, unlocking several practical applications. In this section, we explore two key use cases.
\vspace{-0.3cm}
\paragraph{Concurrent Updates.}
We investigate whether it is possible to merge multiple sets of changes in a single scene, even when the corresponding images contain non-overlapping modifications.
For example, consider one change where a toy bunny is placed on a sofa and another where a computer is removed from a desk, as illustrated in \figref{fig:batched} (a) and (b).
To handle such cases, we apply our method in parallel to each change and afterwards unify the scene, allowing updates as soon as changes are detected. After optimization, we obtain two reconstructions, $\mathcal{G}^t_1$ and $\mathcal{G}^t_2$, corresponding to the full scenes. The sets $\mathcal{O}^t_1$ and $\mathcal{O}^t_2$ represent the localized 3D Gaussians associated with the changed regions. Given these $\mathcal{O}^{t}_1, \mathcal{O}^t_2$ and their indices $\mathcal{I}^t_1, \mathcal{I}^t_2$ in the reconstructions, we can now construct a unifying $\mathcal{G}^t$ by replacing the local regions of the two changes in $\mathcal{G}^{t-1}$. For a detailed explanation of the merging process please refer to the supplementary material.
Our results, depicted in \figref{fig:batched} (c), confirm that merging changes using our method is seamless and practical. Notably, NeRF-based approaches like CL-NeRF~\cite{wu2024cl} cannot merge concurrent changes due to the constraints of their implicit representations. This further highlights the advantage of explicit 3DGS representations, which allow for efficient scene updates without retraining.

\vspace{-0.3cm}
\paragraph{History Recovery.}
\label{sec:history}
Tracking scene history is crucial in 3D reconstruction, as it enables a system to understand environmental changes over time. For a home robot, this means reconstructing scenes from different moments, allowing it to track object movements (e.g., detecting when an item is placed on a table or removed from a shelf). This historical knowledge enhances the robot’s ability to assist users, making it a more intelligent and adaptive system.
Existing methods, such as CL-NeRF\cite{wu2024cl} and Neural Scene Chronology\cite{lin2023neural}, struggle with scene history recovery because they rely on a single global representation, conditioning the model on timestamps to encode scene changes over time. This approach suffers from two fundamental issues: a) limited capacity: compressing all past information into a single model restricts its ability to handle long-term, evolving changes.
b) catastrophic forgetting: the model gradually loses memory of previous states unless changes are repeatedly observed.
In contrast, our method efficiently recovers past 3D reconstructions without requiring large memory overhead or redundant storage.
Given a sequence of reconstructions $(\mathcal{G}^{t})_{t \leq T}$, where we only store the latest state $\mathcal{G}^T$, our goal is to recover any past scene $\mathcal{G}^i$ at a given time step $i \leq T$. A na\"ive approach would be to save every past reconstruction to disk, but this leads to massive storage requirements. Instead, our method leverages shared scene information, requiring storage only for localized changes and their indices in the reconstruction, making it far more memory-efficient.
We evaluated our explicit history recovery method on the real-world portion of the \emph{CL-Splats} dataset and found that it requires only 36MB of additional storage per step, compared to the 1173MB needed for naïve storage on average. Notably, our method fully reconstructs the unchanged scene from previous time steps without any loss in quality.
For a detailed explanation of our algorithm and results from an additional multi-day experiment, we refer readers to the supplementary material.
\vspace{-0.2cm}
\section{Conclusion}
\vspace{-0.2cm}
We introduce \methodname, a framework for continual learning of Gaussian Splats that efficiently adapts to local scene changes. By leveraging dynamic, localized optimization, it updates reconstructions without extensive retraining, making it ideal for large, evolving environments.
\methodname achieves significant speed and reconstruction quality improvements over state-of-the-art methods through precise change detection and targeted optimization. Our experiments demonstrate its ability to adapt efficiently with minimal new data while enabling novel applications such as history recovery and batched updates. These capabilities make \methodname particularly valuable for mixed reality, autonomous navigation, and embodied AI, where maintaining high-quality, up-to-date reconstructions is crucial.
\vspace{-0.8cm}
\paragraph{Limitations.} While our method is robust to various change types and real-world data, it is inherently limited by the base 3D representation’s reconstruction capability. Additionally, since \methodname assumes localized changes, it does not handle global illumination variations. Future research could explore integrating geometry and appearance updates to address this limitation.

\noindent \textbf{Acknowledgements.}
This project was supported by Google, an ARL grant W911NF-21-2-0104 and a Vannevar Bush Faculty Fellowship.
Jonas Kulhanek was supported by Grant Agency of the Czech Technical University in Prague, grant No. SGS24/095/OHK3/2T/13 by the Czech Science Foundation~(GAČR) EXPRO (grant no. 23-07973X).

{
    \small
    \bibliographystyle{ieeenat_fullname}
    \bibliography{main}
}
\pagebreak
\maketitlesupplementary

\definecolor{Best}{HTML}{FF999A}  %
\definecolor{Second}{HTML}{FFCC99}  %
\definecolor{Third}{HTML}{FFFF99}  %

In this \textbf{Supplementary Material}, we provide additional details and discussions to complement the main paper. In \cref{sec:implementation}, we elaborate on the implementation details of the method introduced in Sec.\ 3 of the main paper. \Cref{sec:dataset} describes the datasets used in Sec.\ 4 of the main paper. In \cref{sec:discussions}, we present further insights into the design choices of our method. As highlighted in Sec.\ 5 of the main paper, we demonstrate our method's ability to handle multiple time steps with dynamic changes in \cref{sec:multi}. Finally, \cref{sec:results} includes the full quantitative results of the experiments discussed in Sec.\ 4 of the main paper.
Please also check our supplementary video for more visualizations.

\section{Implementation Details}
\label{sec:implementation}
In our experiments, we utilize the official 3D Gaussian Splatting~\cite{kerbl20233d} implementation as our basis, adopting all default hyperparameters as provided in 3DGS~\cite{kerbl20233d}.

\paragraph{Detecting Changes in 2D.}
We start by obtaining a vanilla 3DGS representation (optimized on time $T_a$) and rendering it from the same viewpoints as the new captures (time $T_b; b>a$).
In our implementation, we use DINOv2 as the feature extractor. In particular, we use the Dinov2-small-14 configuration, which enables real-time processing on an NVIDIA Quadro RTX 6000, facilitating dynamic change detection while exploring a scene. DINOv2 model expects image height and width to be divisible by the patch size 14. To achieve this, we perform a center crop with box width $w_{14}=w-(w\mod{14})$ and height $h_{14}=h-(h\mod{14})$.
After this preprocessing step, we normalize the inputs using the statistics from imagenet~\cite{deng2009imagenet}. The normalized result is fed into DINOv2, and we extract the last layer patch token activations of shape $(\lfloor w/14 \rfloor, \lfloor h/14\rfloor)$. 
We compute the DINOv2 features of the 3DGS $T_a$ rendering and the ground-truth images at $T_b$ and compare them using cosine similarity, giving us a down-scaled changed map. To scale the result to the original resolution, we first resize the change map to $(w_{14}, h_{14})$ using bilinear interpolation and then pad it with $0$s.
Now, we translate this soft mask to a binary mask by thresholding the cosine similarity at $\tau_1=0.5$. Finally, we dilate the mask by 2\% of the image width. We implement this via a convolution with a kernel having only $1$s as weights and then thresholding at $0$.

\paragraph{Identifying Changed 3D Regions via Majority Voting.}
Given the 2D binary masks for each render-capture pair, we now lift these masks into 3D, giving us the set of 3DGS that are affected by the change.
For each frame, we use the camera-projection-matrix to project~\cite{hartley2003multiple} the center of each Gaussian into 2D.
We count the number of times each Gaussian $g_i$ projects into the 2D mask with $c_i$ and how often it projects outside of the image with $o_i$. Based on these counts, we perform a majority voting strategy that independently assigns each Gaussian either to the set of changed Gaussians $\mathcal{O}^{t}$ or to the rest:
\begin{equation}
    g_i \in \mathcal{O}^{t} \quad\text{\textit{iff.}}\quad\frac{4}{3}o_i < |\bI^t| < 2 c_i.
\end{equation}
Here, $\bI^t$ is the set of new images described in Sec.\ 3 of the main paper.
This guarantees that the changed region is well observed while admitting less robust detection in 2D.

Given this set of Gaussians, we split them into clusters $(k_i)$ using HDBSCAN~\cite{mcinnes2017hdbscan} with a minimum cluster size of $1000$.
For each $k_i$, we compute the mean $m_i$ and the $98$th quantile of the Euclidean distance $d_i$ of all points in $k_i$ to $m_i$. We then define a sphere $s_i$ with center $m_i$ and radius $d_i \cdot 1.1$ for each $k_i$.

\paragraph{Sampling New Points.}
As described in the main paper, to guarantee that there will be Gaussians to optimize, we sample points in the area of change. If there are already points close to the area of change, then we can sample in that local area. To do that, we fit the K-Means clustering algorithm on the xyz coordinates of the values with $K=10$ and then reinterpret these as a GaussianMixture and sample new points from there. By sampling few points each round, we gradually expand the 3D region where points are sampled without sampling too many points that will be far away.
In contrast if there are no Gaussians in the area of change, then we start by sampling everywhere.
In any case, we only keep the sampled points that project into enough 2D masks according to the majority voting.
In Alg.~\autoref{alg:sample_full} and Alg.~\autoref{alg:sample_region} we describe how we sample new points across the scene and around existing points. The function \emph{Random} samples points uniformly distributed in the given interval, the function \emph{K-Means} computes a K-Means clustering with $K$ clusters, the function \emph{InitGMMFromClusters} constructs a Gaussians Mixture model with mean being the cluster center and the covariance being the a diagonal matrix containing the vector from the center to the furthest point on its diagonal. Finally \emph{MajorityVote} filters the points according to the voting algorithm described in the main paper. 
While it is a viable approach to always sample in the entire scene, we found that sampling in the region can help to get points faster.

\begin{algorithm}[t]
\caption{Full-Scene Point Sampling}\label{alg:sample_full}
\begin{algorithmic}[1]
    \State \textbf{Input:} $\mathcal{G}^{t-1}$,$(\mathcal{M}_i^t)_{i\leq N}$, $n$
    \State \textbf{Output:} pointcloud $\mathcal{P}$
    \Function{RandomSample}{ $\mathcal{G}^{t-1}$,$(\mathcal{M}_i^t)_{i\leq N}$, $n$}
        \State $min = \min(\mathcal{G}^{t-1}.\text{xyz})$
        \State $max = \max(\mathcal{G}^{t-1}.\text{xyz})$
        \State $\mathcal{S} = $ \Call{Random}{$min$, $max$, $n$}
        \State $\mathcal{S}_{\text{in}} = $ \Call{MajorityVote}{$(\mathcal{M}_i^t)_{i\leq N}$, $\mathcal{S}$}
        \State \Return $\mathcal{S}_{\text{in}}$
    \EndFunction
\end{algorithmic}
\end{algorithm}

\begin{algorithm}[t]
\caption{Region Point Sampling}\label{alg:sample_region}
\begin{algorithmic}[1]
    \State \textbf{Input:} $\mathcal{O}$, $(\mathcal{M}_i^t)_{i\leq N}$, $n$
    \State \textbf{Output:} pointcloud $\mathcal{P}$
    \Function{SampleRegion}{$\mathcal{O}$,$(\mathcal{M}_i^t)_{i\leq N}$, $n$}
        \State $\mathcal{K} =$ \Call{K-Means}{$\mathcal{O}$.\text{xyz}, $K=10$}
        \State $\text{GMM} = $ \Call{InitGMMFromClusters}{$\mathcal{K}$}
        \State $\mathcal{S} = $ \Call{Sample}{$\text{GMM}$, $n/5$}
        \State $\mathcal{S}_{\text{in}} = $ \Call{MajorityVote}{$(\mathcal{M}_i^t)_{i\leq N}$, $\mathcal{S}$}
        \State \Return $\mathcal{S}_{\text{in}}$
    \EndFunction
\end{algorithmic}
\end{algorithm}

\paragraph{3DGS Optimization in Changed Regions.}
We are able to only perform local updates by only applying gradients to Gaussians in $\mathcal{O}^{t}$. We achieve this by masking the gradients for each Gaussian not in that set and only applying optimizer update steps to the Gaussians in this set. This also prevents unwanted movement due to first and second moment from Adam~\cite{kingma2014adam}. In addition to this, we perform pruning of Gaussians in $\mathcal{O}^{t}$ that are not inside any sphere $s_i$ -- every 15 iterations, we check if a Gaussian $g_i\in \bigcup_{j}s_j$ and if not we remove it.
We found that on average less than $1$ Gaussian leaves the spheres each iteration which leads to a smooth optimization.
On the other hand, if they are not pruned then a significant amount of Gaussians will not be in the area of the change making the optimization with the local kernel inefficient.

\paragraph{Local Optimization Kernel.}
We modify the 3DGS CUDA kernel to exploit the spatial locality inherent to our optimization approach. We conducted a preliminary experiment using a real-world scene to justify these modifications. In this experiment, we measured the time spent in the rasterization routines under two conditions: (a) the original complete 3DGS optimization and (b) an artificially constrained version where rendering is restricted to a central crop occupying 30\% of the image.

\tabref{tab:timings} presents the results of these measurements, detailing the timings for individual steps during 3DGS optimization. The results demonstrate that localized optimization achieves speedups proportional to the reduced number of tiles rendered.
\begin{table}[h]
    \centering
    \begin{tabular}{ccc}
        \toprule
         Routing / Method & Full 3DGS & 3DGS 30\% \\
         \midrule
         Forward::Preprocess & 0.70ms & 0.61ms \\
         Forward::Render & 0.92ms & 0.21ms \\
         Backward::Preprocess & 0.32ms & 0.21ms \\
         Backward::Render & 10.99ms & 3.37ms \\
         \midrule
         Total & 12.93ms & 4.40ms \\
         \bottomrule
    \end{tabular}
    \caption{\textbf{Comparison of Individual Routine Times.} We artificially restrict the optimization to a center crop making up 30\% of the tiles and compare the times of the routines.}
    \label{tab:timings}
\end{table}

Given this information, we make three modifications to the diff-gaussian-rasterization kernel:

\begin{enumerate}
    \item We compute the dynamic rendering mask by projecting all Gaussians from $\mathcal{O}^{t}$ onto a boolean matrix of size $\lceil w / 16 \rceil \times \lceil h / 16 \rceil$, setting each entry to \emph{true} if at least one Gaussian projects onto the corresponding tile.
    To avoid additional overhead, we reuse the computed projection to create the per-tile depth ordering.
    \item During forward and backward rendering, we utilize the generated mask to terminate threads associated with tiles marked as \emph{false} in the boolean matrix. We can compute the entry associated with a thread by checking the threats group index for $x$ and $y$.
    \item Finally, we terminate threads performing the backward pass of the preprocessing step for Gaussians that are not part of $\mathcal{O}^{t}$.
\end{enumerate}

\paragraph{Baselines.}
\begin{enumerate}
    \item \textbf{3DGS}~\cite{kerbl20233d}: In our experiments we use the official implementation of 3DGS with unchanged hyperparameters. 
    \item \textbf{GaussianEditor}~\cite{chen2024gaussianeditor}: We used the local optimization code from GaussianEditor~\cite{chen2024gaussianeditor} in combination with our masks. We leave the hyperparameters mostly unchanged -- the only change we make is that we set the number of iterations from $1000$ to $30000$ for a fair comparison.
    \item \textbf{CL-NeRF}~\cite{wu2024cl}: We use the official implementation with the default hyperparameters. Additionally, we make sure that the parameters work robustly on new data as well.
    \item \textbf{CLNeRF}~\cite{cai2023clnerf}: We use the official implementation with the default hyperparameters.
\end{enumerate}

\paragraph{Camera Pose Estimation.} The first step in creating photorealistic reconstructions is obtaining camera positions.
For the synthetic part of the \emph{CL-Splats} dataset, we export the camera positions from Blender, eliminating any influence camera pose estimation has on the reconstruction quality.
For the in-the-wild images, we follow existing methods \cite{wu2024cl, cai2023clnerf} and use COLMAP to recover the camera positions of all the captured images at all times together.
We found that this works well and that COLMAP does not have problems with scene changes encountered in these datasets. We expected this since COLMAP was designed to handle even images from internet collections, which vary in style and geometry from frame to frame and not only between time steps. At the same time, this allows us to conduct our experiments and focus on the reconstruction part; in practice, reconstructing from scratch after each record is not feasible.
COLMAP already includes the necessary functionality to match existing images to a reconstruction.
In particular, the following steps can be taken to add new images to the model.
\begin{enumerate}
    \item Extract features of the new images with \emph{feature\_extractor}.
    \item Match the new images against the existing ones, preferably using \emph{vocab\_tree\_matcher}. This matcher only scales linearly in the number of new images.
    \item Use \emph{image\_registrator} to register the new images to the reconstruction.
\end{enumerate}

The authors of COLMAP suggest using \emph{bundle\_adjuster} to improve the accuracy of the reconstruction, but currently, they only support adjusting over all cameras. A valuable addition for the continual setting would be to allow bundle adjusting only over the new images.

Finally, if too many changes have accumulated over time, it is possible to render the existing 3DGS reconstruction from some viewpoints and use that to start a new reconstruction.

We have experimented with this incremental model and compared it to the jointly estimated model on scenes from the real-world dataset.
We observed that incrementally building the model only led to an average PSNR drop of 1.2 while enabling speedups of up to 20 times.

\section{Details on \emph{CL-Splats} Dataset}
\label{sec:dataset}
In this section, we provide additional details for the datasets that we introduced in Sec.\ 4 of the main paper.
\paragraph{Synthetic Data.}
\begin{figure*}[ht]
    \centering
    \includegraphics[width=\linewidth]{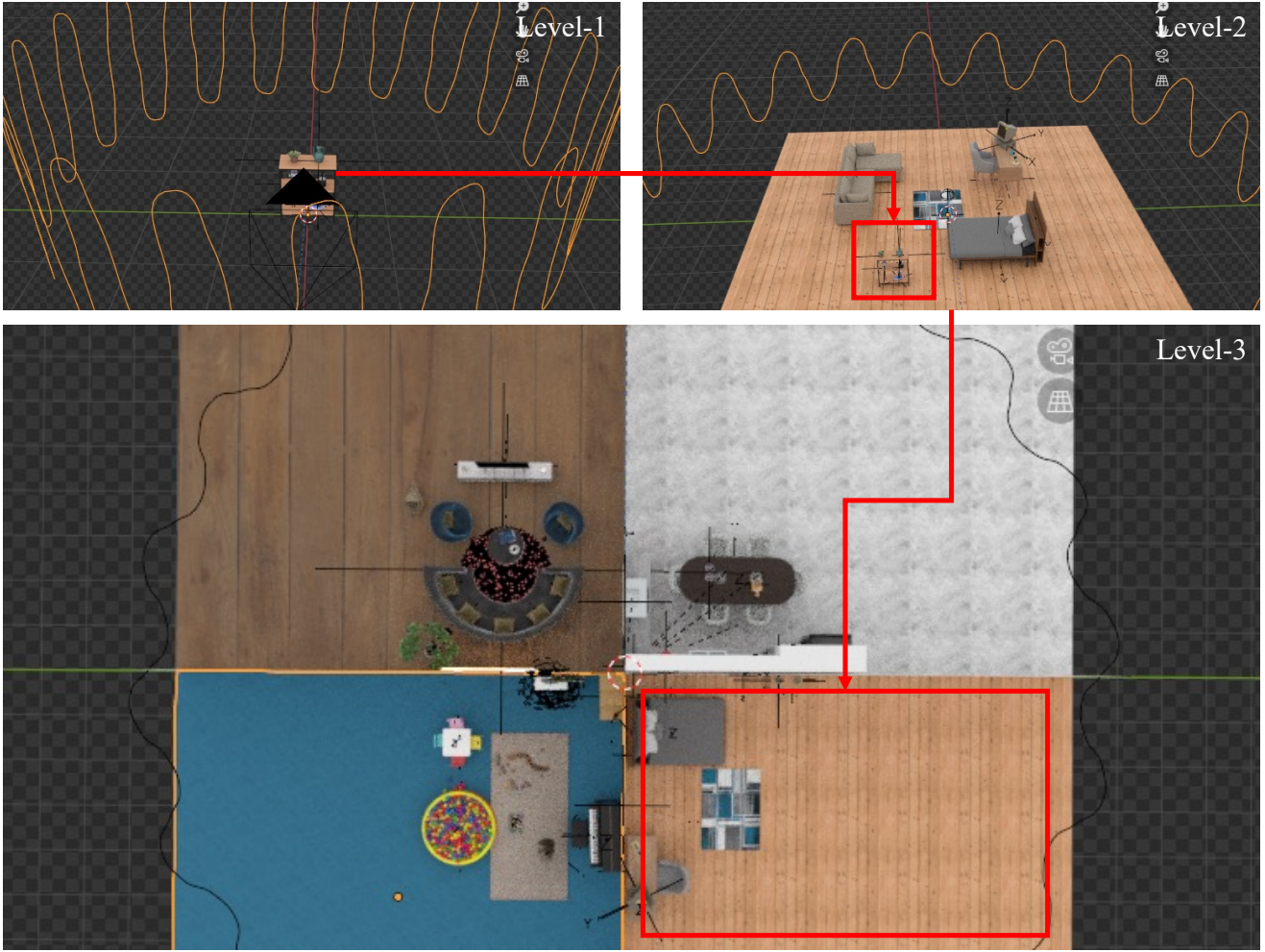}
    \caption{\textbf{Overview of the Synthetic Levels.} Each level is part of the next larger level. The location of the smaller scene is marked by red rectangles in the larger scenes. The curves that are visible in each scene, are the trajectories of the camera for the initial reconstruction.}
    \label{fig:levels}
\end{figure*}
We constructed a synthetic dataset comprising three scenes of varying complexity using Blender, incorporating objects sourced from Objaverse~\cite{deitke2023objaverse} and BlenderKit~\cite{BlenderKit}. Each scene features multiple camera trajectories inspired by Mip-NeRF-360~\cite{barron2022mip}. Notably, the training and test trajectories are designed to be distinct.

\begin{figure*}
    \centering
    \includegraphics[width=\linewidth]{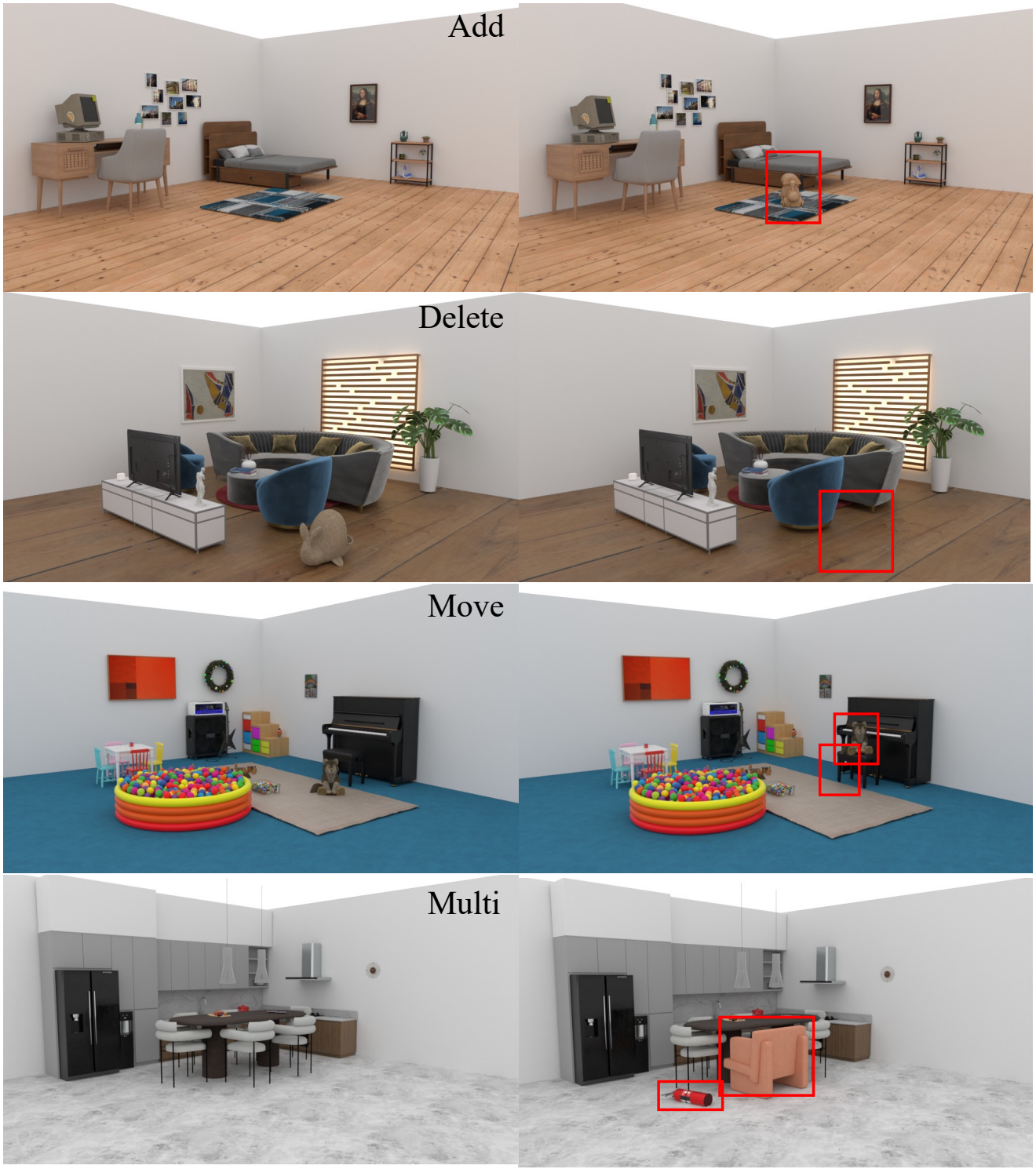}
    \caption{\textbf{Changes in Level-3.} Each column shows the effect of one change on the Level-3 dataset. The changed areas are highlighted by red rectangles. 1) a wooden train is added onto the carpet. 2) The whale basket is removed. 3) The husky plushy gets moved. 4) The white chair gets replaced by a red armchair and a fire extinguisher appears on the floor.}
    \label{fig:changes}
    \vspace{10pt}
\end{figure*}

Each level of complexity incorporates the following types of changes: (1) addition of a new object, (2) removal of an existing object, (3) repositioning of an existing object, and (4) combinations of multiple operations simultaneously. \figref{fig:changes} illustrates the modifications introduced in Level-3.

The levels progressively increase in complexity, with each level encompassing the previous ones, as illustrated in \figref{fig:levels}. The first scene shows a shelf that contains multiple objects. This shelf is found in the bottom left corner of the room-scale level. Finally, this room is part of the Level-3 configuration that is comprised of four different rooms with a large variety of objects and textures.
Below, we outline the key characteristics of each scene.

\begin{enumerate}
    \item \textbf{Level-1:} The area of the scene is $1m^2$. The camera is $3m$ away from the scene center and the focal length of the camera is $50mm$. The changing objects make up 10-20\% of the scene.
    \item \textbf{Level-2:} The area of the scene is $100m^2$. The camera is $5m$ away from the scene center and the focal length of the camera is $50mm$. The changing objects make up $1-2\%$ of the scene.
    \item \textbf{Level-3:} The area of the scene is $400m^2$. The camera is $10m$ away from the scene center and the focal length of the camera is $50mm$. The changing objects make up $<1\%$ of the scene.
\end{enumerate}

For each level, we use 200 frames for the initial reconstruction and 25 for training and testing the changed scene.
Since 3DGS~\cite{kerbl20233d} requires a sparse reconstruction, we generate a point cloud using COLMAP~\cite{schonberger2016structure}. Given that we already have accurate camera poses and intrinsics from Blender, we adopt the pipeline from Tetra-NeRF~\cite{kulhanek2023tetra} to produce a sparse point cloud.

\paragraph{Real-World Data.}
We casually captured five different scenes using an iPhone 13 at 60 FPS. Four of these scenes were recorded indoors, and one was captured outdoors. The outdoor scene and one indoor scene followed Mip-NeRF-360-style trajectories, while the remaining three indoor scenes utilized Zip-NeRF~\cite{barron2023zip}-like trajectories. For each scene, we extract every 20th frame, resulting in 100-200 images for the initial reconstruction and 10-30 images for the changed parts.

\begin{figure}[ht]
    \centering
    \includegraphics[width=\linewidth]{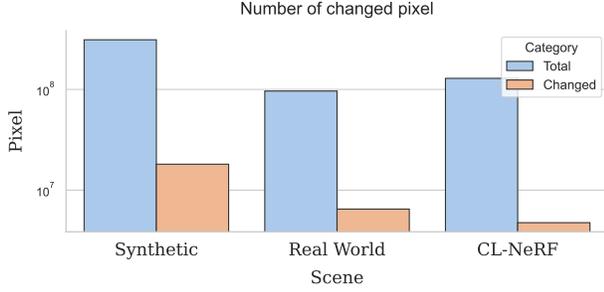}
    \caption{\textbf{Number of Pixels in Changes.} Shows the number of pixels that have changed according to our 2D masks compared to the total number of pixels. The y-axis is in log-scale.}
    \label{fig:pixel-change}
\end{figure}

\begin{figure}[ht]
    \centering
    \includegraphics[width=\linewidth]{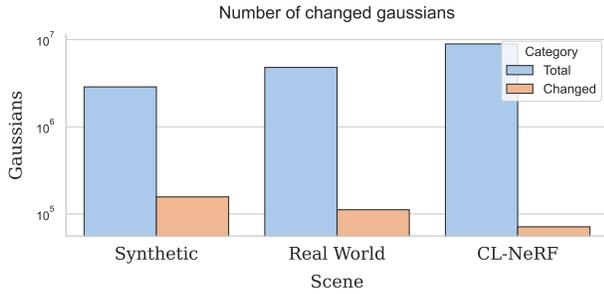}
    \caption{\textbf{Number of Gaussians in Changes.}Shows the number of Gaussians that have changed according to our 3D masks compared to the total number of Gaussians. The y-axis is in log-scale.}
    \label{fig:gaussians-change}
\end{figure}

\paragraph{Statistics.}
In addition to the scene descriptions, we have gathered statistics about the number of pixels and Gaussians constituting the changes in all scenes of all datasets.
We have obtained these numbers by counting the inliers in the 2D masks that we predict as described in \secref{sec:implementation} and the number of Gaussians in $\mathcal{O}^{t}$. \figref{fig:pixel-change} shows the number of pixels that have changed compared to the total number of pixels in each dataset. We can see the variety of change complexities in the datasets. A similar trend can be seen in \figref{fig:gaussians-change} but for the number of Gaussians.

\section{Discussions}
\label{sec:discussions}
\subsection{Local Optimization Without Static Elements}
In GaussianEditor~\cite{chen2024gaussianeditor}, the authors propose rendering only the modified parts of the scenes. Specifically, their renderer considers only the changed Gaussians, entirely omitting the background. While they leverage this approach to refine an existing Gaussian model, we use it to generate new objects from randomly sampled Gaussians. Additionally, their method optimizes for only 1,000 iterations, whereas we optimize for 30,000.

In our main paper, we present the results of their optimization using our masks in Sec.\ 4.1 of the leading paper. However, we also explored how GaussianEditor would perform if tasked with pruning points during optimization. We observed that even for small objects, the optimization failed to converge. It diverged to the point where all points were pruned, causing the optimization to crash.

In contrast, our method works seamlessly with the same set of masks. This highlights the importance of accounting for existing structures during local optimization.

\subsection{Failure Case}
\begin{figure*}[ht]
    \centering
    \includegraphics[width=\linewidth]{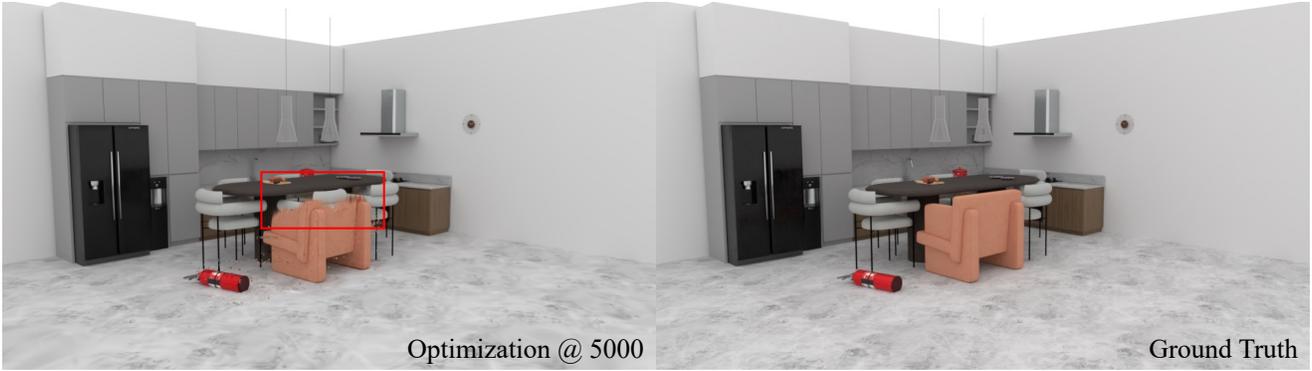}
    \caption{\textbf{Illustration of a Failure Case.} Due to the underestimation of the changed area, the optimization can not possibly make the appropriate changes to the scene. The area of interest is marked by a red rectangle.}
    \label{fig:failure}
\end{figure*}

As discussed in Sec.\ 4.2, achieving a high recall in the changed area is critical for reconstruction quality. Existing structures cannot be effectively optimized without sufficient recall, and new structures are constrained too tightly. \figref{fig:failure} illustrates an example where the estimated area is too small, resulting in an inability to modify the scene during optimization properly.
We observed that this happens when some structures are extremely thin. Despite DinoV2 being a semantic model, we found that we can reliably segment changes even when objects are replaced by new ones with similar semantics.
We also experimented with na\"ive sampling methods which a) samples a fixed number of points in the entire scene or b) samples a fixed number of points in a fixed radius of the inliers and found that these lead to up to 6~PSNR lower scores.

\section{Batched Updates}
To obtain a unified reconstruction $\mathcal{G}^t$ that reflects both sets of changes, we track the indices of the modified Gaussians, $\mathcal{I}^{t-1}_1$ and $\mathcal{I}^{t-1}_2$, in the original representation. The final scene is constructed by combining the updated Gaussians $\mathcal{O}^t_1$ and $\mathcal{O}^t_2$ with the unchanged Gaussians from the original scene:

\begin{equation} \mathcal{G}^t = \mathcal{O}^{t}_1 \cup \mathcal{O}^{t}_2 \cup \mathcal{G}^{t-1} \setminus \mathcal{G}^{t-1}[\overline{\mathcal{I}^{t-1}_1} \cap \overline{\mathcal{I}^{t-1}_2}]. \end{equation}

For ease of notation, we treat Gaussians and indices as indexable sets. This formulation ensures that all unchanged Gaussians are retained, while the updated ones are seamlessly integrated.

By leveraging this approach, our method successfully merges scene updates, enabling efficient reconstruction of complex dynamic environments without the need for full re-optimization.

\section{Multi Day}

\subsection{History Recovery}
\paragraph{Algorithm.} To efficiently recover any past reconstruction $ \mathcal{G}^i $, we introduce an algorithm that stores only the changing Gaussians $ \mathcal{O}^t $ and their corresponding indices $ \mathcal{I}^t $. This allows us to reconstruct past states without storing redundant information. We assume that in $ \mathcal{G}^{t} $, the first  

\begin{equation}
\#^{t}_\text{static} := |\mathcal{G}^{t-1} \setminus \mathcal{O}^t|
\end{equation}

Gaussians correspond to the static parts of the previous state:

\begin{equation}
\mathcal{G}_{\text{static}}^{t-1} := \mathcal{G}^{t-1} \setminus \mathcal{O}^t.
\end{equation}

To satisfy this requirement, we store $\mathcal{O}^t$ and $\mathcal{I}^t$ immediately after computing $\mathcal{O}^t $, before any optimization. Additionally, before optimization, we rearrange the Gaussian order to ensure:

\begin{equation}
\mathcal{G}^{t}[:\#^{t}_\text{static}] = \mathcal{G}_{\text{static}}^{t-1}.
\end{equation}

Since duplication and pruning only affect $ \mathcal{O}^t $, this condition remains invariant throughout optimization. To reconstruct $ \mathcal{G}^{t-1} $ from $ \mathcal{G}^{t} $, we:
Extract $\mathcal{G}_{\text{static}}^{t-1} $ by accessing the first:

\begin{equation}
    \#^{t}_\text{static} = |\{i \mid \forall i \in \mathcal{I}^t, i=0\}|.
\end{equation}

2. Retrieve the stored $ \mathcal{O}^t $ and reconstruct the full scene as:

\begin{equation}
    \mathcal{G}^{t-1} := \mathcal{O}^t \cup \mathcal{G}_{\text{static}}^{t-1}.
\end{equation}

3. Re-arrange the Gaussians to maintain consistency:
$ \mathcal{G}_{\text{static}}^{t-1} $ is placed where $ \mathcal{I}^t = 0 $ and changing Gaussians $ \mathcal{O}^t $ are placed where $ \mathcal{I}^t = 1 $. Alg.~\autoref{alg:example} shows how this process can be repeated to recover $ \mathcal{G}^k $ from $ \mathcal{G}^t $, enabling multi-step history recovery. Our algorithm allows for efficient and exact scene recovery while drastically reducing storage requirements. As shown in the main paper, it achieves identical reconstructions to storing the full scene while using significantly less memory.

\begin{algorithm}[t]
\caption{History Recovery}\label{alg:example}
\begin{algorithmic}[1]
    \State \textbf{Input:} $\mathcal{G}^T$, $(\mathcal{O}^k)_k$,$(\mathcal{I}^k)_{k}$, $n$
    \State \textbf{Output:} $\mathcal{G}^{n}$
    \Function{RecoverState}{$\mathcal{G}^T$,$(\mathcal{O}^k)_{k}$, $(\mathcal{I}^k)_{k}$, $n$}
        \If{$n = T$}
            \State \Return $\mathcal{G}^T$
        \Else
            \State $\mathcal{N} = \mathcal{G}^T[:|\neg\mathcal{I}^T|] + \mathcal{O^T}$ \Comment{Array of target size}
            \State $\mathcal{N}[\mathcal{I}^T] = \mathcal{O}^T$ \Comment{Set indices to dynamic}
            \State $\mathcal{N}[\neg\mathcal{I}^T] = \mathcal{C}$ \Comment{Set complement to static}
            \State \Return \Call{RecoverState}{$\mathcal{N}$, $(\mathcal{O}^k)_{k}$, $(\mathcal{I}^k)_{k}$, $n$}
        \EndIf
    \EndFunction
\end{algorithmic}
\end{algorithm}

\subsection{Multi Day Reconstruction}
\label{sec:multi}
In Sec.\ 5 of the main paper, we discussed the shortcomings of existing methods to scale to multiple days and how our method can elegantly solve this problem due to the explicit representation of the changes.
This section showcases our method's prowess over multiple days (4). As a basis, we use the second level of our blender dataset.

\begin{figure*}[ht]
    \centering
    \includegraphics[width=\linewidth]{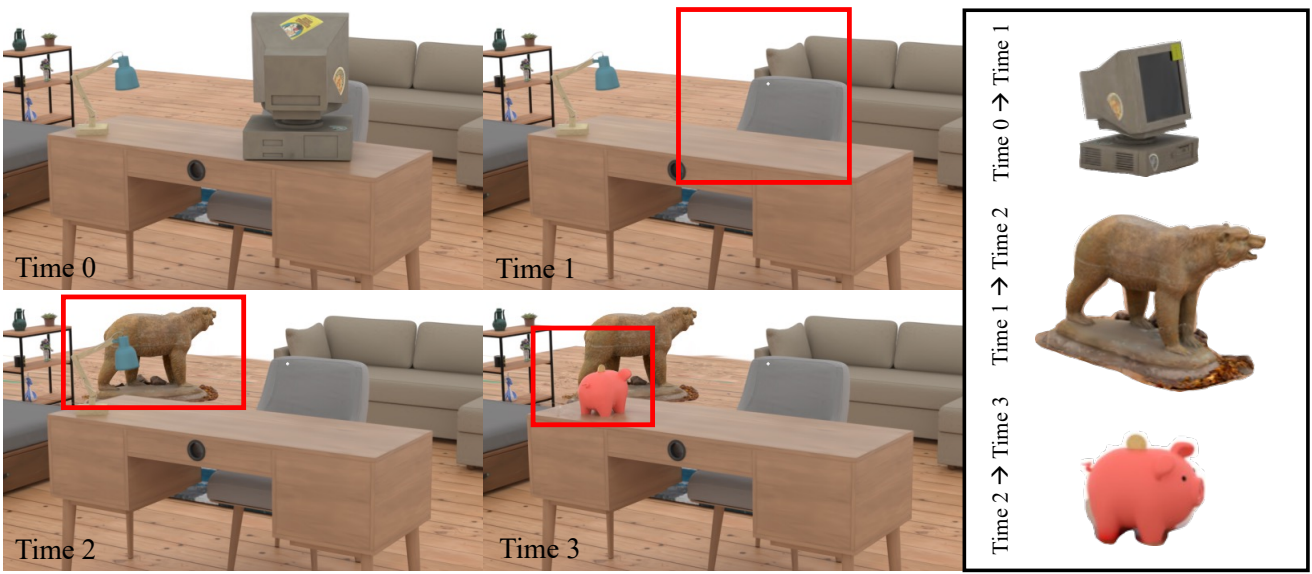}
    \caption{\textbf{Reconstruction Results over Multiple Days.} In this figure we show the same scene and viewpoint at four different points in time. The changed region is highlighted in red. On the right side are the changed objects that can be used to efficiently recover any point in time. From 0 to 1, the computer on the desk gets removed. From 1 to 2, a statue of a bear gets added in the back. From 2 to 3, the desk lamp gets replaced by a piggy bank.}
    \label{fig:multi}
\end{figure*}

\paragraph{Setup.}
We perform three different operations leading to four reconstructions based on our Level-2 synthetic dataset. For each change, we provide 25 images for training focusing on the changed region.

\paragraph{Results.}
\figref{fig:multi} shows some qualitative results of the reconstructions in each time step. We can see that unchanged regions stay the same over the time steps and that no error accumulates due to forgetting. Additionally, the figure separately shows the objects involved in the change. Our method has the advantage that saving a new day's reconstruction is as simple as saving the Gaussians of the changed object and their original indices. With just the base scene and the changed objects, we can go back in time, reconstructing any time steps scene immediately and at no loss in quality.

\section{Complete Results}
\label{sec:results}
In this section, we present the extended results to the CL-Splats dataset and the CL-NeRF dataset from Sec.\ 4.1.

\subsection{\emph{CL-Splats} Dataset}
\paragraph{Level-1.}
\Cref{tab:level-1} presents the per-operation performance of all methods on our Level-1 dataset. Our method demonstrates a clear advantage, outperforming all baselines substantially. Among the baseline methods, 3DGS shows competitive performance in this scenario, benefiting from the relatively high scene coverage compared to other levels. However, the local optimization of 3DGS within 2D masks (3DGS+M) proves to be particularly challenging, leading to a noticeable drop in overall performance despite the simplicity of the setting.

GaussianEditor, which also employs local optimization, demonstrates clear limitations due to its reliance on 2D masks and the absence of background consideration during optimization. This drawback is particularly evident in the PSNR scores achieved during object deletion, where GaussianEditor performs noticeably worse than other methods. Unlike other approaches that can effectively handle object removal, GaussianEditor's failure to account for the background significantly impacts its metrics.

Additionally, we observed that CL-NeRF struggles to converge on this dataset. Even when the training duration was doubled to 400,000 steps, no significant improvement in results was achieved. This limitation appears to stem from the inherent characteristics of the underlying NeRF representation.

\paragraph{Level-2.}
As shown in \Cref{tab:level-2}, the performance of all methods is evaluated on the Level-2 dataset. Compared to Level-1, the baseline methods—3DGS, 3DGS+M, and GaussianEditor—demonstrate a noticeable decline in performance, indicating increased difficulty at this level. In contrast, our method remains highly consistent, underscoring its robustness across varying levels of complexity. Interestingly, the challenges associated with object deletion also persist at this level.

CL-NeRF achieves improved reconstruction quality compared to Level-1, though it still lags significantly behind our method.

\paragraph{Level-3.}
We can see in \tabref{tab:level-3} that
Level-3 continues to follow the established trends, supporting our synthetic data design. The baseline methods show a further decline in performance compared to Level-2, while our method delivers stable and reliable results. Interestingly, CL-NeRF exhibits improved performance on this dataset, likely due to the greater object distances, which make PSNR less sensitive to the smoother reconstructions it generates.

\paragraph{Real-World Data.}
\Cref{tab:real-life} presents the complete results on real-world data. While the performance gap between our method and the baseline approaches narrows in this setting, our method outperforms the others on average. Unlike CL-NeRF and GaussianEditor, our approach demonstrates reliability, avoiding catastrophic failures.

\subsection{CL-NeRF Dataset}

\paragraph{Whiteroom.}
As shown in \Cref{tab:whiteroom}, despite the Whiteroom's sparse textures and challenging lighting conditions, our method achieves performance comparable to CL-NeRF. In contrast, the baseline 3DGS performs significantly worse, mainly due to the limited coverage of the scene's unchanged regions.

Additionally, we observe that move-and-replace operations result in poorer reconstruction quality than add-and-delete operations. While this can be partly attributed to the inherent complexity of these operations, the primary challenge stems from the placement of the objects involved. These objects are often situated in distant corners, where reduced view coverage adversely impacts reconstruction accuracy.

\paragraph{Kitchen.}
\Cref{tab:kitchen} provides the complete results for the Kitchen scene from the CL-NeRF dataset. On this dataset, our method achieves results comparable to those of CL-NeRF. However, CL-NeRF faces significant challenges in accurately reconstructing the scene after object deletion, movement, or replacement operations. In contrast, our approach produces reliable reconstructions, demonstrating its robustness.

\paragraph{Rome.}
\Cref{tab:rome} highlights our method's ability to reconstruct large-scale scenes with remarkable accuracy and ease, such as the Colosseum in Rome. Comparatively, CL-NeRF faces significant challenges in delivering detailed reconstructions for such complex scenarios.

\begin{table}[t!]
\centering
\scriptsize %
\begin{tabular}{lcccccc}
\toprule
\textbf{Method/Ops} & \multicolumn{3}{c}{\textbf{Add}} & \multicolumn{3}{c}{\textbf{Delete}} \\
\cmidrule(lr){2-4} \cmidrule(lr){5-7}
    & PSNR$\uparrow$ & LPIPS$\downarrow$ & SSIM$\uparrow$ & PSNR$\uparrow$ & LPIPS$\downarrow$ & SSIM$\uparrow$ \\
\midrule
3DGS~\cite{kerbl20233d} & 10.688 & 0.445 & 0.313 & 6.697 & 0.499 & 0.345 \\
CL-NeRF~\cite{wu2024cl} & \cellcolor{Second}26.416 & \cellcolor{Second}0.123 & \cellcolor{Second}0.807 & \cellcolor{Second}23.364 & \cellcolor{Second}0.154 & \cellcolor{Second}0.761 \\
\textbf{Ours} & \cellcolor{Best}34.474 & \cellcolor{Best}0.009 & \cellcolor{Best}0.979 & \cellcolor{Best}34.961 & \cellcolor{Best}0.009 & \cellcolor{Best}0.979 \\
\bottomrule
\end{tabular}
\caption{\textbf{Reconstruction Quality on the Rome Dataset from the CL-NeRF dataset.}}
\label{tab:rome}
\end{table}
\begin{table*}[ht]
\centering
\scriptsize
\begin{tabular}{lcccccccccccc}
\toprule
\textbf{Method/Operation} & \multicolumn{3}{c}{\textbf{Add}} & \multicolumn{3}{c}{\textbf{Delete}} & \multicolumn{3}{c}{\textbf{Move}} & \multicolumn{3}{c}{\textbf{Multi}} \\
\cmidrule(lr){2-4} \cmidrule(lr){5-7} \cmidrule(lr){8-10} \cmidrule(lr){11-13}
    & PSNR$\uparrow$ & LPIPS$\downarrow$ & SSIM$\uparrow$ & PSNR$\uparrow$ & LPIPS$\downarrow$ & SSIM$\uparrow$ & PSNR$\uparrow$ & LPIPS$\downarrow$ & SSIM$\uparrow$ & PSNR$\uparrow$ & LPIPS$\downarrow$ & SSIM$\uparrow$ \\
\midrule
3DGS~\cite{kerbl20233d} & 23.018 & \cellcolor{Third}0.051 & \cellcolor{Third}0.956 & \cellcolor{Third}27.707 & \cellcolor{Second}0.021 & \cellcolor{Second}0.977 & \cellcolor{Second}25.897 & \cellcolor{Second}0.030 & \cellcolor{Second}0.972 & \cellcolor{Second}25.947 & \cellcolor{Second}0.030 & \cellcolor{Second}0.972\\
3DGS+M & 18.802 & 0.101 & 0.899 & 18.669 & 0.108 & 0.894 & 18.927 & 0.116 & 0.889 & 19.682 & 0.096 & 0.907 \\
GaussianEditor~\cite{chen2024gaussianeditor} & \cellcolor{Third}25.548 & \cellcolor{Second}0.047 & \cellcolor{Second}0.973 & 22.863 & 0.065 & \cellcolor{Third}0.957 & 24.193 & \cellcolor{Third}0.051 & \cellcolor{Third}0.959 & \cellcolor{Third}25.254 & \cellcolor{Third}0.044 & \cellcolor{Third}0.967\\
CL-NeRF~\cite{wu2024cl} & \cellcolor{Second}26.541 & 0.051 & 0.947 & \cellcolor{Second}26.918 & \cellcolor{Third} 0.045 & 0.953 & \cellcolor{Third}24.902 & 0.053 & 0.946 & 25.203 & 0.054 & 0.947\\
\textbf{CL-Splats (ours)} & \cellcolor{Best}41.923 & \cellcolor{Best}0.001 & \cellcolor{Best}0.998 & \cellcolor{Best}44.018 & 
\cellcolor{Best}0.001 & \cellcolor{Best}0.998 & \cellcolor{Best}37.481 & \cellcolor{Best}0.007 & \cellcolor{Best}0.994 & \cellcolor{Best}36.310 & \cellcolor{Best}0.007 & \cellcolor{Best}0.994 \\
\bottomrule
\end{tabular}
\caption{\textbf{Reconstruction Quality on Level-1 of the Synthetic Data from the CL-Splats Dataset.}}
\label{tab:level-1}
\end{table*}
\begin{table*}[ht]
\centering
\scriptsize
\begin{tabular}{lcccccccccccc}
\toprule
\textbf{Method/Operation} & \multicolumn{3}{c}{\textbf{Add}} & \multicolumn{3}{c}{\textbf{Delete}} & \multicolumn{3}{c}{\textbf{Move}} & \multicolumn{3}{c}{\textbf{Multi}} \\
\cmidrule(lr){2-4} \cmidrule(lr){5-7} \cmidrule(lr){8-10} \cmidrule(lr){11-13}
    & PSNR$\uparrow$ & LPIPS$\downarrow$ & SSIM$\uparrow$ & PSNR$\uparrow$ & LPIPS$\downarrow$ & SSIM$\uparrow$ & PSNR$\uparrow$ & LPIPS$\downarrow$ & SSIM$\uparrow$ & PSNR$\uparrow$ & LPIPS$\downarrow$ & SSIM$\uparrow$ \\
\midrule
3DGS~\cite{kerbl20233d} & 17.319 & 0.245 & 0.761 & \cellcolor{Third}26.081 & \cellcolor{Third}0.125 & 0.854 & \cellcolor{Third}26.496 & 0.115 & 0.874 & 25.674 & 0.113 & 0.881 \\
3DGS+M & 13.731 & 0.330 & 0.715 & 14.118 & 0.322 & 0.639 & 14.845 & 0.339 & 0.679 & 13.581 & 0.382 & 0.635 \\
GaussianEditor~\cite{chen2024gaussianeditor} & \cellcolor{Third}22.386 & \cellcolor{Third}0.109 & \cellcolor{Third}0.926 & 19.521 & 0.162 & \cellcolor{Third}0.887 & 24.792 & \cellcolor{Third}0.072 & \cellcolor{Third} 0.942 & \cellcolor{Third} 26.913 & \cellcolor{Third} 0.074 & \cellcolor{Third}0.939 \\
CL-NeRF~\cite{wu2024cl} & \cellcolor{Second}28.845 & \cellcolor{Second}0.077 & \cellcolor{Second}0.918 & \cellcolor{Second}31.157 & \cellcolor{Second}0.068 & \cellcolor{Second}0.928 & \cellcolor{Second}31.076 & \cellcolor{Second}0.070 & \cellcolor{Second}0.927 & \cellcolor{Second}28.134 & \cellcolor{Second}0.109 & \cellcolor{Second}0.894 \\
\textbf{CL-Splats (ours)} & \cellcolor{Best}39.528 & \cellcolor{Best}0.025 & \cellcolor{Best}0.978 & \cellcolor{Best}41.647 & \cellcolor{Best}0.016 & \cellcolor{Best}0.982 & \cellcolor{Best}41.838 & \cellcolor{Best}0.017 & \cellcolor{Best}0.9815 & \cellcolor{Best}40.329 & \cellcolor{Best}0.015 & \cellcolor{Best}0.979 \\
\bottomrule
\end{tabular}
\caption{\textbf{Reconstruction Quality on Level-2 of the Synthetic Data from the CL-Splats Dataset.}}
\label{tab:level-2}
\end{table*}
\begin{table*}[ht]
\centering
\scriptsize
\begin{tabular}{lcccccccccccc}
\toprule
\textbf{Method/Operation} & \multicolumn{3}{c}{\textbf{Add}} & \multicolumn{3}{c}{\textbf{Delete}} & \multicolumn{3}{c}{\textbf{Move}} & \multicolumn{3}{c}{\textbf{Multi}} \\
\cmidrule(lr){2-4} \cmidrule(lr){5-7} \cmidrule(lr){8-10} \cmidrule(lr){11-13}
    & PSNR$\uparrow$ & LPIPS$\downarrow$ & SSIM$\uparrow$ & PSNR$\uparrow$ & LPIPS$\downarrow$ & SSIM$\uparrow$ & PSNR$\uparrow$ & LPIPS$\downarrow$ & SSIM$\uparrow$ & PSNR$\uparrow$ & LPIPS$\downarrow$ & SSIM$\uparrow$ \\
\midrule
3DGS~\cite{kerbl20233d} & \cellcolor{Third}16.509 & 0.389 & 0.696 & \cellcolor{Third}15.602 & \cellcolor{Third}0.395 & 0.678 & \cellcolor{Third}18.010 & \cellcolor{Third}0.349 & \cellcolor{Third}0.756 & \cellcolor{Third}15.653 & \cellcolor{Third}0.402 & 0.680 \\
3DGS+M & 12.638 & 0.437 & 0.682 & 12.397 & 0.403 & 0.693 & 12.767 & 0.505 & 0.601 & 11.366 & 0.500 & 0.611 \\
GaussianEditor~\cite{chen2024gaussianeditor} & 15.581 & \cellcolor{Third}0.378 & \cellcolor{Third}0.725 & 10.443 & 0.434 & \cellcolor{Third}0.732 & 10.951 & 0.481 & 0.718 & 9.191 & 0.446 & \cellcolor{Third}0.718 \\
CL-NeRF~\cite{wu2024cl} & \cellcolor{Second}34.560 & \cellcolor{Second}0.043 & \cellcolor{Second}0.951 & \cellcolor{Second}34.326 & \cellcolor{Second}0.045 & \cellcolor{Second}0.952 & \cellcolor{Second}34.772 & \cellcolor{Second}0.045 & \cellcolor{Second}0.951 & \cellcolor{Second}34.326 & \cellcolor{Second}0.045 & \cellcolor{Second}0.952 \\
\textbf{CL-Splats (ours)} & \cellcolor{Best}39.902 & \cellcolor{Best}0.023 & \cellcolor{Best}0.979 & \cellcolor{Best}39.842 & \cellcolor{Best}0.024 & \cellcolor{Best}0.979 & \cellcolor{Best}40.027 & \cellcolor{Best}0.023 & \cellcolor{Best}0.979 & \cellcolor{Best}38.669 & \cellcolor{Best}0.025 & \cellcolor{Best}0.978 \\
\bottomrule
\end{tabular}
\caption{\textbf{Reconstruction Quality on Level-3 of the Synthetic Data from the CL-Splats Dataset.}}
\label{tab:level-3}
\end{table*}
\begin{table*}[ht]
\setlength{\tabcolsep}{3pt}
\centering
\tiny %
\resizebox{\textwidth}{!}{%
\begin{tabular}{lccccccccccccccccccccc}
\toprule
\textbf{Method/Operation} & \multicolumn{3}{c}{\textbf{Cone}} & \multicolumn{3}{c}{\textbf{Shoe}} & \multicolumn{3}{c}{\textbf{Shelf}} & \multicolumn{3}{c}{\textbf{Room}} & \multicolumn{3}{c}{\textbf{Desk}}\\
\cmidrule(lr){2-4} \cmidrule(lr){5-7} \cmidrule(lr){8-10} \cmidrule(lr){11-13} \cmidrule(lr){14-16}
    & PSNR$\uparrow$ & LPIPS$\downarrow$ & SSIM$\uparrow$ & PSNR$\uparrow$ & LPIPS$\downarrow$ & SSIM$\uparrow$ & PSNR$\uparrow$ & LPIPS$\downarrow$ & SSIM$\uparrow$ & PSNR$\uparrow$ & LPIPS$\downarrow$ & SSIM$\uparrow$ & PSNR$\uparrow$ & LPIPS$\downarrow$ & SSIM$\uparrow$ \\
\midrule
3DGS~\cite{kerbl20233d} & 13.449 & 0.471 & 0.136 & 6.592 & 0.409 & 0.326 & \cellcolor{Third}20.211 & \cellcolor{Third}0.120 & \cellcolor{Third}0.860 & 10.061 & \cellcolor{Third}0.471 & 0.326 & 8.506 & 0.409 & 0.345 \\
3DGS+M & 8.900 & 0.578 & 0.117 & 5.976 & 0.462 & 0.226 & 12.177 & 0.496 & 0.599 & 8.359 & 0.385 & 0.184 & 7.515 & 0.381 & 0.226 \\
GaussianEditor~\cite{chen2024gaussianeditor} & \cellcolor{Second} 26.277 & \cellcolor{Second} 0.140 & \cellcolor{Second}0.850 & \cellcolor{Second}25.969 & \cellcolor{Second}0.053 & \cellcolor{Second}0.921 & 10.211 & 0.434 & 0.671 & \cellcolor{Second}27.944 & \cellcolor{Second}0.039 & \cellcolor{Second}0.950 & \cellcolor{Best}30.268 & \cellcolor{Best}0.050 & \cellcolor{Best}0.944\\
CL-NeRF~\cite{wu2024cl} & \cellcolor{Third} 21.482 & \cellcolor{Third}0.422 & \cellcolor{Third}0.511 & \cellcolor{Third} 25.583 & \cellcolor{Third}0.192 & \cellcolor{Third}0.829 & \cellcolor{Best} 25.731 & \cellcolor{Second}0.154 & \cellcolor{Second}0.844 & \cellcolor{Third}15.125 & 0.551 & \cellcolor{Third}0.574 & \cellcolor{Third}28.419 & \cellcolor{Third}0.132 & \cellcolor{Third}0.867 \\
\textbf{Ours} & \cellcolor{Best}27.163 & \cellcolor{Best}0.114 & \cellcolor{Best}0.886 & \cellcolor{Best} 28.144 & \cellcolor{Best}0.046 & \cellcolor{Best}0.926 & \cellcolor{Second}25.256 & \cellcolor{Best}0.064 & \cellcolor{Best}0.944 & \cellcolor{Best}31.203 & \cellcolor{Best}0.0375 & \cellcolor{Best}0.965 & \cellcolor{Second}29.477 & \cellcolor{Second}0.066 & \cellcolor{Second}0.929 \\
\bottomrule
\end{tabular}%
}
\caption{\textbf{Reconstruction Quality on Real-World Data from the CL-Splats Dataset.}}
\label{tab:real-life}
\end{table*}
\begin{table*}[ht]
\centering
\scriptsize %
\begin{tabular}{lcccccccccccc}
\toprule
\textbf{Method/Ops} & \multicolumn{3}{c}{\textbf{Add}} & \multicolumn{3}{c}{\textbf{Delete}} & \multicolumn{3}{c}{\textbf{Move}} & \multicolumn{3}{c}{\textbf{Replace}} \\
\cmidrule(lr){2-4} \cmidrule(lr){5-7} \cmidrule(lr){8-10} \cmidrule(lr){11-13}
    & PSNR$\uparrow$ & LPIPS$\downarrow$ & SSIM$\uparrow$ & PSNR$\uparrow$ & LPIPS$\downarrow$ & SSIM$\uparrow$ & PSNR$\uparrow$ & LPIPS$\downarrow$ & SSIM$\uparrow$ & PSNR$\uparrow$ & LPIPS$\downarrow$ & SSIM$\uparrow$ \\
\midrule
3DGS~\cite{kerbl20233d} & 11.503 & 0.275 & 0.733 & 14.839 & 0.233 & 0.773 & 11.717 & 0.284 & 0.731 & 11.524 & 0.282 & 0.728 \\
CL-NeRF~\cite{wu2024cl} & \cellcolor{Second}31.736 & \cellcolor{Second}0.056 & \cellcolor{Second}0.935 & \cellcolor{Best}34.021 & \cellcolor{Best}0.053 & \cellcolor{Best}0.938 & \cellcolor{Best}29.121 & \cellcolor{Second}0.089 & \cellcolor{Best}0.906 & \cellcolor{Best}29.917 & \cellcolor{Second}0.082 & \cellcolor{Best}0.912 \\
\textbf{Ours} & \cellcolor{Best}34.014 & \cellcolor{Best}0.053 & \cellcolor{Best}0.953 & \cellcolor{Second}31.821 & \cellcolor{Second}0.068 & \cellcolor{Second}0.935 & \cellcolor{Second}27.889 & \cellcolor{Best}0.078 & \cellcolor{Second}0.909 & \cellcolor{Second}27.805 & \cellcolor{Best}0.078 & \cellcolor{Second}0.909 \\
\bottomrule
\end{tabular}
\caption{\textbf{Reconstruction Quality on Whiteroom from the CL-NeRF dataset.}}
\label{tab:whiteroom}
\end{table*}
\begin{table*}[ht]
\centering
\scriptsize %
\begin{tabular}{lcccccccccccc}
\toprule
\textbf{Method/Ops} & \multicolumn{3}{c}{\textbf{Add}} & \multicolumn{3}{c}{\textbf{Delete}} & \multicolumn{3}{c}{\textbf{Move}} & \multicolumn{3}{c}{\textbf{Replace}} \\
\cmidrule(lr){2-4} \cmidrule(lr){5-7} \cmidrule(lr){8-10} \cmidrule(lr){11-13}
    & PSNR$\uparrow$ & LPIPS$\downarrow$ & SSIM$\uparrow$ & PSNR$\uparrow$ & LPIPS$\downarrow$ & SSIM$\uparrow$ & PSNR$\uparrow$ & LPIPS$\downarrow$ & SSIM$\uparrow$ & PSNR$\uparrow$ & LPIPS$\downarrow$ & SSIM$\uparrow$ \\
\midrule
3DGS~\cite{kerbl20233d} & 19.486 & 0.4021 & 0.511 & 8.487 & 0.602 & 0.4074 & 7.339 & 0.5573 & 0.4025 & 8.444 & 0.583 & 0.425 \\
CL-NeRF~\cite{wu2024cl} & \cellcolor{Second}27.199 & \cellcolor{Best}0.263 & \cellcolor{Best}0.789 & \cellcolor{Second}24.290 & \cellcolor{Best}0.316 & \cellcolor{Best}0.751 & \cellcolor{Second}24.610 & \cellcolor{Best}0.304 & \cellcolor{Best}0.759 & \cellcolor{Second}23.241 & \cellcolor{Second}0.334 & \cellcolor{Best}0.736 \\
\textbf{Ours} & \cellcolor{Best}27.546 & \cellcolor{Second}0.296 & \cellcolor{Second}0.695 & \cellcolor{Best}26.927 & \cellcolor{Second}0.323 & \cellcolor{Second}0.6738 & \cellcolor{Best}27.291 & \cellcolor{Second}0.319 & \cellcolor{Second}0.681 & \cellcolor{Best}27.151 & \cellcolor{Best}0.322 & \cellcolor{Second}0.679 \\
\bottomrule
\end{tabular}
\caption{\textbf{Reconstruction Quality on Kitchen from the CL-NeRF dataset.}}
\label{tab:kitchen}
\end{table*}

\immediate\closein\imgstream

\end{document}